%% file: main-preprint.tex
\documentclass[11pt]{article}

\IfFileExists{acl.sty}{%
  \usepackage[preprint]{acl}
}{%
  \usepackage[a4paper,margin=2.5cm,twocolumn,columnsep=0.6cm]{geometry}%
  \usepackage{natbib}%
  \setcitestyle{authoryear,round}%
  \usepackage[colorlinks=true,allcolors=blue]{hyperref}%
}

\usepackage{times}
\usepackage{latexsym}
\usepackage[T1]{fontenc}
\usepackage[utf8]{inputenc}
\usepackage{microtype}
\usepackage{inconsolata}
\usepackage{graphicx}
\usepackage{booktabs}
\usepackage{multirow}
\usepackage{amsmath,amssymb}
\usepackage{xcolor}
\usepackage{enumitem}
\usepackage{tabularx}
\usepackage{xltabular}   % page-breaking, page-width taxonomy table (Appendix A)
\usepackage{colortbl}    % \rowcolor for its group header rows
\usepackage{arydshln}    % \cdashline (dashed rules in Table 1); load after colortbl
\definecolor{basecolor}{gray}{0.90}
\usepackage{verbatim}
\usepackage{tikz}         % overview figure (Figure 1)
\usetikzlibrary{positioning,arrows.meta,fit,backgrounds,calc}
\usepackage{pifont}       % \ding symbols
\DeclareUnicodeCharacter{2500}{-}
\DeclareUnicodeCharacter{2014}{--}
\DeclareUnicodeCharacter{00D7}{x}

\graphicspath{{figures/}}

\input{macros}

\input{codestyle}   % code listings style

\title{LogicVal:A Logic-Formalized Benchmark for Constraint Satisfaction in Text-to-Image Generation}

\author{
  Savitha Sam Abraham\textsuperscript{1,*,\dag},
  Marjan Alirezaie\textsuperscript{2,3,*,\dag},
  Hossein Rahnama\textsuperscript{2,3,4} \\
  \textsuperscript{1}City St George's, University of London \\
  \textsuperscript{2}Toronto Metropolitan University \quad
  \textsuperscript{3}Flybits Labs, Creative AI Hub, Toronto \\
  \textsuperscript{4}MIT Media Lab, Massachusetts Institute of Technology \\
  {\small 
  \texttt{savitha.sam-abraham@city.ac.uk}, \texttt{marjan.alirezaie@torontomu.ca}, \texttt{rahnama@mit.edu}}
}

\begin{document}
\maketitle

\renewcommand{\thefootnote}{}
\footnotetext{* Equal contribution. \dag\ Corresponding authors.}
\renewcommand{\thefootnote}{\arabic{footnote}}

\input{sections/abstract}
\input{sections/introduction}

\input{sections/related_work}

\input{sections/logicval}

\input{sections/experimental_setup}
\input{sections/results}
\input{sections/metric_reliability}

\input{sections/conclusion}

\input{sections/limitations}

\input{sections/ethics}

\bibliography{references}
\IfFileExists{acl_natbib.bst}{\bibliographystyle{acl_natbib}}{\bibliographystyle{plainnat}}

\clearpage
\appendix
\input{sections/appendix/background_program}
\input{sections/appendix/constraint_templates}
\input{sections/appendix/prompts}
\input{sections/appendix/benchmark_statistics}
\input{sections/appendix/verifier_and_metrics}

\input{sections/appendix/lora_training}
\input{sections/appendix/human_annotation}
\input{sections/appendix/additional_results}

\end{document}

%% file: macros.tex
\newcommand{\bench}{LogicVal}
\newcommand{\gpt}{GPT-Image-2}
\newcommand{\gemini}{Gemini-3-Pro-Image}
\newcommand{\flux}{FLUX.1-dev}
\newcommand{\qwen}{Qwen-Image}
\newcommand{\clingo}{\textsc{clingo}}

\newcommand{\pkg}[1]{\textsc{#1}}

\newcommand{\Sgold}{S^{*}}

\newcommand{\repolink}
{\href{https://github.com/ccig-tutorial/logicval}
{LogicVal-repo}}

%% file: codestyle.tex
\usepackage{listings}
\usepackage[most]{tcolorbox}
\lstdefinelanguage{ASP}{morecomment=[l]{\%}}
\newtcblisting{codebox}[1][]{
  listing only, breakable, enhanced,
  colback=black!4, colframe=black!35, boxrule=0.5pt, arc=2pt,
  left=4pt, right=4pt, top=2pt, bottom=2pt, before skip=6pt, after skip=8pt,
  listing options={
    language=ASP, basicstyle=\ttfamily\footnotesize, columns=fixed, basewidth=0.5em,
    keepspaces=true, upquote=true, showstringspaces=false, breaklines=true,
    commentstyle=\color{black!55}, #1
  },
}

%% file: sections/abstract.tex
\begin{abstract}
Text-to-Image (T2I) generative models have made significant progress in producing high-fidelity images from natural-language prompts. Alongside this progress, numerous benchmarks have been developed to evaluate capabilities such as attribute binding, numeracy, spatial reasoning, and commonsense reasoning. In this work, we propose a new perspective on T2I evaluation by casting it as a constraint-satisfaction problem. We introduce a logical framework that formally specifies the benchmark domain, including objects, attributes, and relations, alongside the constraints that generated images must satisfy. We evaluated four state-of-the-art T2I models across 9 classes of constraints ranging from a single-object-attribute to multi-object relationships constraints involving logical operators ($\forall, \exists, count, \rightarrow$). Our results reveal a substantial gap between single-object and complex conditional multi-object relationship  constraint classes, with the best-performing model achieving 98\% accuracy on the former and just 53\% on the latter. We assess compositional generalisation, finding that the best-performing model`s accuracy drops from 84\% on single-constraint prompts to 20\% when satisfying multiple constraints simultaneously. Identifying unsatisfiable constraints is also challenging, with the best model correctly recognizing only 23\% of such prompts. Building on this framework, we introduce \textbf{LogicVal}, an evaluation methodology that combines neural visual detection with logical reasoning to assess constraint satisfaction in generated images. On CLEVR, LogicVal achieves ROC-AUC scores ranging from 0.88 to 0.98, indicating strong alignment with human judgments. These findings underscore the value of logic-based evaluation for characterizing T2I model capabilities and limitations.
\end{abstract}

%% file: sections/introduction.tex
\section{Introduction}
\label{sec:intro}
Text-to-Image (T2I) generative models have made impressive progress in recent years, producing images that capture both realistic scenarios and highly imaginative concepts. This progress has enabled the widespread adoption of T2I models across a range of applications including fashion, architecture \citep{li2026dit, ni2025ai}, and beyond. 
As T2I models become increasingly capable, an important question arises: What comes next? A key challenge is moving beyond generating visually compelling images toward generating images that reliably satisfy the constraints specified in a prompt. Current T2I models typically treat prompts as a form of soft conditioning, providing no formal guarantee that all specified requirements will be satisfied in the generated image. This gap is particularly important for applications where controllability and reliability are critical. The appeal and educational potential of T2I models in medical pedagogy have been investigated by \citet{temsah2024art}, which highlights concerns regarding their reliability in educational settings.  

Enabling further progress toward controllable and reliable T2I generation requires benchmarks that can systematically characterize what a model can and cannot satisfy. Existing T2I benchmarks have made important progress in this direction, with much of the literature focusing on three core capabilities: attribute binding to objects, numeracy, and spatial relationships between objects \citep{huang2025t2i}. These benchmarks typically define their own sets of objects, attributes, and relations, together with prompt templates that are instantiated using elements from these sets. Evaluation is then commonly performed using multimodal large language model (MLLM)-based metrics like Soft-TIFA \citep{kamath2025geneval} or other learned similarity measures like CLIPScore \citep{hessel2021clipscore}. 
Existing T2I evaluation often lacks a formal specification of the properties that an image is required to satisfy. We introduce such a specification, and use it to turn evaluation into a logical constraint-satisfaction problem.

\input{sections/fig_overview}

In this paper, we propose a new perspective on T2I benchmarking by placing logical formalization at its core. We realize this by formally specifying two key components: (i) the \textbf{\textit{domain specification}}, which defines the objects, attributes, and relations considered in the benchmark, and (ii) the \textbf{\textit{constraint specification}}, which defines the constraints on properties and relationships that a generated image is required to satisfy (see Figure~\ref{fig:overview}). Both specifications are expressed in Answer Set Programming (ASP) \citep{eiter2009answer}; the domain specification is domain-agnostic, and we instantiate it here for CLEVR \citep{johnson2017clevr}. The constraint specification supports templates ranging from simple attribute binding to more complex relational structures involving multiple objects and relations, including graph-structured compositions. This formalization decouples the specification of the evaluation criteria from their natural-language realization and, importantly, casts T2I evaluation as a constraint-satisfaction problem that can be addressed by combining neural detection models for visual grounding with logical solvers for constraint verification.
%Building on this framework, we introduce LogicVal, a logic-based evaluation methodology that combines symbolic reasoning with neural visual detection models. Given a generated image and a set of formally specified constraints, LogicVal first grounds relevant visual concepts using neural detection models \cite{ren2024grounded} and subsequently uses logical reasoning to determine whether the grounded image satisfies the specified constraints. This provides a more explicit and interpretable evaluation of T2I models than metrics based solely on learned similarity or free-form MLLM judgments. Moreover, the logical representation provides a natural way to characterize model capabilities: when developing and evaluating T2I models, one can formally specify the classes of constraints that a model is expected to satisfy and report performance across these classes.
Building on this formalization, we introduce \textbf{\bench{}}, a benchmark and evaluation framework for logical constraint satisfaction in T2I generation. The contributions of this paper are as follows:

\begin{itemize}
  \item \textbf{The \bench{} benchmark.}\footnote{The code is available at \repolink{}; the benchmark, the human annotations and the generated images will be released with the camera-ready version.} We formalise evaluation criteria as domain specification (objects, attributes, relations) and constraint specification in ASP, decoupling what is evaluated from how it is phrased. We define 9 constraint classes, with each class having 4 to 11 templates of varying complexity and logical expressivity, from attribute binding to quantifier alternation and count comparison, and instantiate them into 509 solver-labelled (SAT/UNSAT) specifications.
  \item \textbf{The \bench{} score and its meta-evaluation.}   We introduce the \bench{} score, which makes the correctness of a T2I-generated image a matter of logical entailment.
  We combine neural grounding with a logic solver to verify an image's alignment with the constraints specified. \bench{} judgement is validated against human-judgement for a set of 3887 images: it reaches a ROC-AUC of 0.88--0.98 with human judgments.
  
  \item \textbf{An analysis of state-of-the-art T2I models.} 
  We evaluate four T2I models on their ability to generate images that satisfy \textbf{\textit{single-constraint}} mentioned in a prompt vs \textbf{\textit{multiple-constraints}} specified jointly in a prompt, as well as their ability to identify prompts containing unsatisfiable constraints. We provide a detailed analysis of the challenges posed by different classes of constraints, individual logical operators ($\forall$, $\exists$, $count$, $\rightarrow$), and their combinations, investigating whether particular constraint types or compositions pose consistent difficulties across models.
\end{itemize}

%% file: sections/fig_overview.tex
\begin{figure*}[t]
\centering
\resizebox{\textwidth}{!}{%
\begin{tikzpicture}[
  font=\sffamily\fontsize{6.8pt}{7.8pt}\selectfont,
  every node/.append style={execute at begin node={\hyphenpenalty=10000\exhyphenpenalty=10000}},
  proc/.style={rectangle, rounded corners=3pt, draw=black!75, thick, align=center, fill=white,
               inner sep=3.5pt, minimum height=1.05cm},
  data/.style={rectangle, rounded corners=3pt, draw=black!75, dashed, align=center, fill=white,
               inner sep=3.5pt},
  arr/.style={-{Stealth[length=2mm]}, thick, black!75},
  lbl/.style={font=\sffamily\fontsize{6.4pt}{7.4pt}\selectfont, black!70, align=center},
  facts/.style={data, align=left, font=\ttfamily\fontsize{5.6pt}{6.6pt}\selectfont, text width=2.9cm},
  stage/.style={rectangle, rounded corners=6pt, inner sep=6pt},
  stagelbl/.style={font=\sffamily\small\bfseries, anchor=south},
  group/.style={rectangle, rounded corners=4pt, draw=black!55, densely dotted, thick, inner sep=4pt},
]
\newcommand{\hd}[1]{{\scriptsize\textbf{#1}}}
\definecolor{cblue}{RGB}{40,90,190}
\definecolor{cgreen}{RGB}{30,130,60}
\definecolor{cred}{RGB}{200,40,40}
% column x-positions (west edges)
\def\xA{0}\def\xB{3.95}\def\xC{7.75}\def\xD{12.45}\def\xE{17.35}\def\xF{21.45}\def\xG{25.3}

% ================================================================ top tier: processes
\node[data, text width=3.0cm, anchor=north west] (dom) at (\xA,0)
  {\hd{Domain specification}\\ objects, attributes, relations;\\ e.g.\ CLEVR (used here), COCO};
\node[proc, text width=3.3cm, anchor=north west] (inst) at (\xB,0)
  {\hd{Instantiation + \textsc{clingo}}\\ fill templates with domain values,\\ solve for $n$ objects:
   \textcolor{cgreen}{\textbf{SAT}} / \textcolor{cred}{\textbf{UNSAT}}};
\node[proc, text width=3.4cm, anchor=north west] (verb) at (\xC,0)
  {\hd{Verbaliser}\\ renders $C$ as a short / long\\ natural-language prompt};
\node[proc, text width=3.3cm, anchor=north west] (t2i) at (\xD,0)
  {\hd{Text-to-image model}\\ GPT-Image-2\\ Gemini-3-Pro-Image\\ FLUX.1-dev\\ Qwen-Image};
\node[proc, text width=3.3cm, anchor=north west] (det) at (\xE,0)
  {\hd{Detection}\\ GSAM + CLIP: shape, colour,\\ material and bounding box\\ of each object};
\node[proc, text width=3.1cm, anchor=north west] (sp) at (\xF,0)
  {\hd{Spatial grounding}\\ regions and spatial relations\\ from the bounding boxes};
\node[proc, text width=1.9cm, anchor=north west] (check) at (\xG,0)
  {\hd{\textsc{clingo}}\\ $\widehat{S}(I)\models C$\,?};

% baseline of the bottom tier, below the tallest process box
\coordinate (bt) at ($(current bounding box.south)+(0,-0.5)$);

% ================================================================ bottom tier: outputs
% C6 template from ccig-dataset-gen/src/eval_dataset_gen/constraint_templates/C6_1prop.txt
% (predicate names shortened: cex = c6_1_prop_counterexample, sat = c6_1_prop_satisfied)
\node[facts, text width=3.0cm, anchor=north west] (cons) at (\xA,0 |- bt)
  {\sffamily\centering\hd{Constraint specification}\\ 9 classes C1--C9,\\ 71 ASP templates (Table~\ref{tab:taxonomy});\\ e.g.\ C6:\par\vspace{2pt}
   \raggedright\ttfamily
   cex(X) :- object(X), object(Y),\\
   \hspace*{1em}X!=Y, hasProperty(X,P1',\textcolor{cgreen}{V1'}),\\
   \hspace*{1em}hasProperty(Y,P2',\textcolor{cblue}{V2'}),\\
   \hspace*{1em}not hasRelationship(X,Y,\textcolor{cred}{D1'}),\\
   \hspace*{1em}\textcolor{cgreen}{V1'}!=\textcolor{cblue}{V2'}.\\
   sat :- object(X), not cex(X),\\
   \hspace*{1em}hasProperty(X,P1',\textcolor{cgreen}{V1'}).\\
   :- not sat.};
% the C6 template above, instantiated for this example (1SAT id 201): P1'=P2'=color,
% V1'=green, V2'=blue, D1'=right; last rule = anti-vacuity (at least one blue object)
\node[facts, text width=3.3cm, anchor=north west] (spec) at (\xB,0 |- bt)
  {\sffamily\centering\hd{Specification $C$}\\ C6, $n{=}5$, \textcolor{cgreen}{\textbf{SAT}}\par\vspace{2pt}
   \raggedright\ttfamily
   cex(X) :- object(X), object(Y),\\
   \hspace*{1em}X!=Y, hasProperty(X,color,\textcolor{cgreen}{green}),\\
   \hspace*{1em}hasProperty(Y,color,\textcolor{cblue}{blue}),\\
   \hspace*{1em}not hasRelationship(X,Y,\textcolor{cred}{right}).\\
   sat :- object(X), not cex(X),\\
   \hspace*{1em}hasProperty(X,color,\textcolor{cgreen}{green}).\\
   :- not sat.\\
   :- \#count\{Y: object(Y),\\
   \hspace*{1em}hasProperty(Y,color,\textcolor{cblue}{blue})\}=0.};
\node[data, text width=3.4cm, align=left, anchor=north west] (prompt) at (\xC,0 |- bt)
  {\hd{Prompt $P_C$} (short)\\[1pt]
   \fontsize{5.9pt}{6.9pt}\selectfont\itshape ``\ldots\ The scene is with 5 objects. There must be at least one \textcolor{cblue}{\textbf{blue}} object in the scene. There exists some \textcolor{cgreen}{\textbf{green}} object such that it is to the \textcolor{cred}{\textbf{right}} of every \textcolor{cblue}{\textbf{blue}} object.''};
\node[inner sep=0pt, draw=black!40, anchor=north west] (img) at (\xD,0 |- bt)
  {\includegraphics[width=3.4cm]{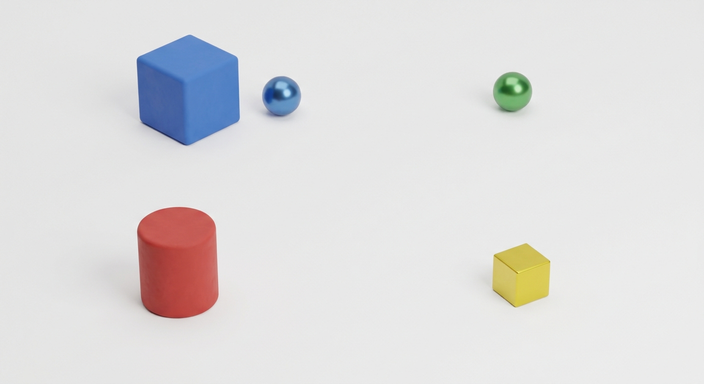}};
\node[lbl, anchor=north] (imglbl) at (img.south) {Gemini-3-Pro-Image};
\node[inner sep=0pt, draw=black!40, anchor=north west] (detimg) at (\xE,0 |- bt)
  {\includegraphics[width=3.4cm]{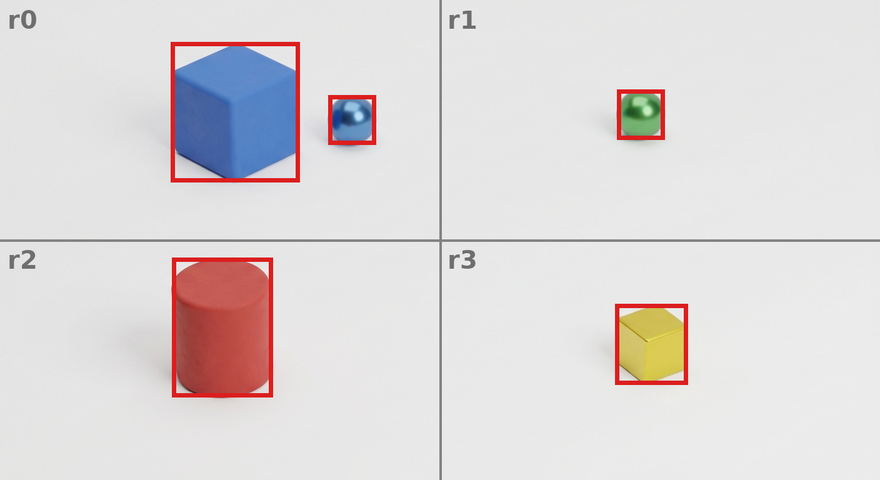}};
\node[lbl, anchor=north] (detlbl) at (detimg.south) {detected bounding boxes and $2\times2$ regions};
\node[facts, text width=3.0cm, anchor=north west] (sg) at (\xF,0 |- bt)
  {\sffamily\hd{Scene graph $\widehat{S}(I)$}\\[1pt]
   object(1). object(2). \ldots\\
   hasProperty(1,color,\textcolor{cblue}{blue}).\\
   hasProperty(2,color,\textcolor{cgreen}{green}).\\
   hasRelationship(2,1,\textcolor{cred}{right}).};
\node[data, text width=1.9cm, anchor=north west] (verdict) at (\xG,0 |- bt)
  {\hd{\bench{} verdict}\\[2.5pt] \textcolor{cgreen}{\ding{51}} satisfied};

% ================================================================ arrows
% generation
\draw[arr] (dom) -- (cons);
\draw[arr] (cons.east) -- ++(0.3,0) |- (inst.west);
\draw[arr] (inst) -- (spec);
\draw[arr] (spec.east) -- ++(0.2,0) |- (verb.west);
\draw[arr] (verb) -- (prompt);
\draw[arr] (prompt.east) -- ++(0.5,0) |- (t2i.west);
% image generation
\draw[arr] (t2i) -- (t2i.south |- img.north) node[lbl, midway, right, inner sep=2pt] {image $I$};
% verification
\draw[arr] (img.east) -- ++(0.8,0) |- (det.west);
\draw[arr] (det) -- (detimg);
\draw[arr] (detimg.east) -- ++(0.2,0) |- (sp.west);
\draw[arr] (detimg.east) -- (detimg.east -| sg.west);
\draw[arr] (sp) -- (sg);
\draw[arr] (sg.east) -- ++(0.2,0) |- (check.west);
\draw[arr] (check) -- (verdict);
% the specification C also goes to the solver (routed below everything)
\coordinate (low) at ($(current bounding box.south)+(0,-0.3)$);
\draw[arr] (spec.south) -- (spec.south |- low) -| ($(check.east)+(0.35,0)$) -- (check.east);
\node[lbl, anchor=south] at ($(check.south |- low)+(-1.6,0)$) {specification $C$};

% ================================================================ backgrounds (shared top line for all stages)
\coordinate (stagetop) at ($(dom.north)+(0,0.45)$);
\begin{scope}[on background layer]
  \node[stage, fill=cblue!6, fit={([xshift=-4pt,yshift=0.35cm]dom.north west)([xshift=-4pt,yshift=-4pt]cons.south west)(inst)(spec)(verb)(prompt)(\xA,0 |- stagetop)}] (s1) {};
  \node[stagelbl] at (s1.north) {\ding{172}\ \bench{} benchmark};
  \node[group, fit={(dom)(cons)}, label={[font=\sffamily\scriptsize\bfseries, anchor=south, inner sep=1pt]north:Formal specification}] (fs) {};
  \node[stage, fill=black!5, fit={(t2i)(img)(imglbl)(\xD,0 |- stagetop)}] (s2) {};
  \node[stagelbl] at (s2.north) {\ding{173}\ Image generation};
  \node[stage, fill=cgreen!6, fit={(det)(detimg)(detlbl)(sp)(sg)(check)(verdict)(\xE,0 |- stagetop)}] (s3) {};
  \node[stagelbl] at (s3.north) {\ding{174}\ \bench{} evaluation};
\end{scope}
\end{tikzpicture}}
\caption{Overview of \bench{}, following one specification (class C6). \ding{172}~The benchmark: a formal specification (domain and constraint templates) is instantiated and labelled SAT/UNSAT by \clingo{} and verbalised into a prompt. \ding{173}~Image generation: the prompt is rendered by a T2I model. \ding{174}~LogicVal evaluation: scene graph is extracted and checked against the constraint specification with \clingo{}.}
\label{fig:overview}
\end{figure*}

%% file: sections/related_work.tex
\input{tables/related_work_comparison.tex}

\section{Related Work}
\label{sec:related}
\subsection{T2I Image Generative Models}
Text-to-Image (T2I) generation has evolved through a variety of generative modeling paradigms, including Generative Adversarial Networks (GANs) \cite{goodfellow2014generative}, Vector-Quantized Variational Autoencoders (VQ-VAEs) \cite{van2017neural}, diffusion models \cite{ho2020denoising,song2020denoising}, normalizing flows \cite{rezende2015variational, chen2018neural}, and, more recently, flow-matching models \cite{lipman2023flow}. 
%esser2024scaling,peebles2023scalable
%Variational Autoencoders (VAEs) \cite{kingma2016improved}, 
%We evaluate models representing recent T2I paradigms: \textbf{Qwen-Image} \cite{wu2025qwen} (diffusion), \textbf{FLUX.1-dev} \cite{bfl2024flux} (flow matching), \textbf{GPT Image} \cite{openai2026gptimage2} (autoregressive generation), and \textbf{Gemini} \cite{google2025geminiimage}, a proprietary multimodal model whose image-generation architecture is not publicly specified in sufficient detail for precise classification.
%\cite{}, 
In this work, we evaluate models spanning multiple recent T2I generation paradigms. Specifically, we evaluate \textbf{Qwen-Image} \cite{wu2025qwen}, which is based on diffusion, and \textbf{FLUX.1-dev}, which uses flow matching, as well as \textbf{GPT Image}, which follows an autoregressive generation paradigm. We additionally evaluate \textbf{Gemini}, a proprietary multimodal model with image-generation capabilities whose underlying image-generation architecture is not publicly specified in sufficient detail for a precise architectural classification. %Our selection therefore allows us to examine constraint satisfaction across different generative paradigms, as well as across open-weight and closed models.

\subsection{T2I Benchmarking}
Existing T2I benchmarks fall into two broad categories: those evaluating literal prompt–image alignment and those assessing reasoning based on implicit commonsense or domain knowledge. The first includes GenEval \cite{ghosh2023geneval}, GenEval 2 \cite{kamath2025geneval}, and T2I-CompBench \cite{huang2023t2icompbench}, which assess attribute binding, object relationships, numeracy, and compositional generation. NegT2IBench \cite{elfatairy2026negt2ibench} focuses on negation, while GenAI-Bench \cite{li2024genaibench} also evaluates counting and universal quantification. These benchmarks primarily use predefined prompt templates, except GenAI-Bench, which uses human-curated prompts.

Table \ref{tab:related} compares these benchmarks with LogicVal. Partial coverage ($\circ$) indicates limited support relative to LogicVal's logical expressiveness and compositional complexity. For example, while existing benchmarks typically test numeracy by checking specified object counts, LogicVal distinguishes exact, minimum, and maximum cardinalities and supports comparative constraints (e.g., equal numbers of cylinders and cubes). It also extends quantifier evaluation beyond simple existential and universal statements to combinations such as $\forall\bar{x}\exists\bar{y}$ and $\exists\bar{x}\forall\bar{y}$, as well as conditional constraints.

The second category includes reasoning-oriented benchmarks such as T2I-ReasonBench \cite{sun2025t2ireasonbench}, R2I-Bench \cite{chen2025r2ibench}, WorldGenBench \cite{zhang2025worldgenbench}, which evaluate whether generated images reflect implicit world knowledge or scientific principles. This knowledge-intensive reasoning lies beyond LogicVal's scope, which focuses on formally specified, explicit logical constraints.

\subsection{T2I Evaluation Metrics}
One challenge in text-to-image (T2I) research is developing automatic evaluation metrics that align closely with human judgments of prompt adherence. We focus on reference-free metrics, including CLIPScore \cite{hessel2021clipscore}, which measures image–prompt alignment using the cosine similarity between their CLIP embeddings. More recent approaches leverage vision-language models (VLMs) for evaluation. VQAScore \cite{lin2024vqascore} computes the probability that a visual question answering (VQA) model answers \textit{Yes} to the question \textit{`Does this image align with the prompt?'} given the generated image. Soft-TIFA \cite{kamath2025geneval} decomposes the prompt into $n$ subquestions, each with an expected answer, and evaluates alignment by aggregating the probabilities assigned to the expected answers by a VQA model.

%% file: tables/related_work_comparison.tex
\begin{table*}[t]
\centering
\tiny
\setlength{\tabcolsep}{6pt}
\newcommand{\y}{\ding{51}}
\newcommand{\n}{--}
\newcommand{\p}{$\circ$}
\begin{tabular}{@{}lcccccc@{\hspace{14pt}}c@{}}
\toprule
\textbf{Benchmark} & \textbf{Negation} & \textbf{$\forall/\exists$} & \textbf{Counting} & \textbf{Composition} & \textbf{UNSAT} & \textbf{Formal spec.}  \\
\midrule
T2I-CompBench \citeyearpar{huang2023t2icompbench} & \n & \p & \p & \y & \n & \n \\
GenEval \citeyearpar{ghosh2023geneval}           & \n & \p & \p & \p & \n & \n  \\
GenEval~2 \citeyearpar{kamath2025geneval}        & \n & \p & \p & \y & \n & \n  \\
GenAI-Bench \citeyearpar{li2024genaibench}       & \y & \p & \y & \y & \n & \n  \\
GeckoNum \citeyearpar{kajic2024geckonum}         & \p & \p & \p & \n & \n & \n  \\
% VERIFY Neg./Hum.
NegT2IBench \citeyearpar{elfatairy2026negt2ibench} & \y & \p & \n & \p & \n & \n  \\
\midrule
\bench{} (ours) & \y & \y & \y & \y & \y & \y  \\
\bottomrule
\end{tabular}
\caption{Comparison with other T2I benchmarks based on logical operators evaluated, presence of unsatisfiable prompts, a formal specification. \ding{51}~covered, $\circ$~partly, --~not covered.}
\label{tab:related}
\end{table*}

%% file: sections/logicval.tex
\section{\bench{}}
\label{sec:logicval}

As shown in Figure~\ref{fig:overview}, \bench{} consists of a benchmark, which turns a formal specification into prompts, and a logic-based evaluation metric, which checks whether the generated images entail the constraint specification.
We first fix the notation (\S\ref{sec:formalization}), then describe the two parts (\S\ref{sec:benchmark}, \S\ref{sec:verification}).

\input{tables/constraint_taxonomy.tex}

% =============================================================================
\subsection{Formalization}
\label{sec:formalization}

A specification $C$ is a set of constraints over the domain; a verbaliser renders it as a prompt $P_C$, and a T2I model $G$ generates an image $I = G(P_C)$.
The \bench{} evaluation extracts a scene graph $\widehat{S}(I)$, i.e.\ the objects, their attributes and their spatial relations, and an ASP solver checks whether $\widehat{S}(I) \models C$.
Since many different images can satisfy the same $C$, no reference image is needed.
If the constraints in $C$ contradict each other, no scene can satisfy them, and the correct output is an empty scene.

% =============================================================================
\subsection{Benchmark}
\label{sec:benchmark}

\paragraph{Domain specification.}
The domain specification defines objects, their attributes, and inter-object relationships on which constraints can be specified. We represent the CLEVR domain \citep{johnson2017clevr}, with its vocabulary of eight colours, three shapes, and two materials, and divide scenes into four regions on a 2×2 grid, following CLEVR-POC \citep{abraham2024clevrpoc}. Spatial relations \emph{left}, \emph{right}, \emph{front}, and \emph{behind} are defined by a background ASP program. Objects in different regions inherit their regions' spatial relations, while objects within the same region are freely ordered subject to anti-symmetry, inverse, and transitivity axioms (Appendix~\ref{app:background}).

\paragraph{Constraint specification.}
\label{sec:taxonomy}
Attribute values and spatial relations are unary and binary predicates, e.g.\ $\mathit{Blue}(x)$ and $\mathit{Left}(x,y)$, both possibly negated.
For objects $\bar{x}=x_1,\dots,x_t$, $\Phi(\bar{x})$ and $\Psi(\bar{x})$ are conjunctions of attribute literals and $R(\bar{x})$ and $R'(\bar{x})$ are conjunctions of spatial relations among them; $R(\bar{x},\bar{y})$ relates $\bar{x}$ to a second group $\bar{y}=y_1,\dots,y_m$, and $\theta\in\{=,\neq,<,\le,>,\ge\}$ compares counts.
Table~\ref{tab:taxonomy} lists nine primitive classes in three groups of increasing expressivity:
\textbf{Object-level} classes (C1--C3) constrain single objects existentially, universally or conditionally;
\textbf{SubGraph-level} classes (C4--C7) constrain several related objects, from existential (C4) and conditional (C5) patterns to the quantifier alternations $\exists\forall$ (C6) and $\forall\exists$ (C7);
\textbf{Cardinality-level} classes compare the number of objects satisfying a pattern with a constant (C8) or with another pattern (C9).
%Appendix~\ref{app:taxonomy} gives the full formal definitions.
%\paragraph{Constraint variants and composition.}
%\label{sec:variants}
We use the logical programming language ASP to realize the above constraint formalization. LogicVal has 71 constraint templates in ASP which are instantiated to generate prompts. Appendix~\ref{app:templates} provides the list of 71 templates. For templates involving universal quantification (e.g., \textit{All blue objects are cylinders}), we add an anti-vacuity clause requiring at least one object to satisfy the antecedent (e.g., \textit{There is at least one blue object}), preventing the statement from being vacuously true. A compositional specification combines $m\in\{1,3,6\}$ constraints from distinct classes, which are jointly checked by the solver.

\paragraph{Instantiation and labelling.}
A constraint specification $C$ is generated by instantiating a template from one of the classes and sampling a scene size $n \in \{3,\dots,9\}$, representing the number of objects. The resulting prompt is: \textit{Generate an image with $n$ objects such that it satisfies constraint $C$.} Given $n$ and $C$, \clingo{} \citep{gebser2019clingo} checks whether a valid configuration of the $n$ objects satisfying $C$ exists. The specification is labelled \pkg{SAT} if an answer set exists and \pkg{UNSAT} otherwise. Appendix~\ref{app:background} provides details of the ASP program supplied to \clingo{}.

\paragraph{Verbalisation.}
A deterministic verbaliser renders each constraint in a \emph{short} and a \emph{long} register; the long one spells out the quantifier scope explicitly.
It also produces sub-questions with expected answers, which are used by the Soft-TIFA metric (\S\ref{sec:metrics-def}). Appendix~\ref{app:prompts} shows the prompt template and examples of prompts in both registers.

\paragraph{Benchmark composition.}
We generate 6 main dataset packages. Each package is a collection of prompts. A package is named by its number of constraints m followed by its label, e.g. 1SAT or 3SAT or 1UNSAT - m denotes the number of constraints to be satisfied in a single prompt. The \bench{} benchmark contains 509 prompts: 355/25 1SAT/1UNSAT, 25/10 3SAT/3UNSAT, and 25/69 6SAT/6UNSAT prompts.

\pkg{1SAT} covers all templates equally (Refer Appendix~\ref{app:bench-stats} for benchmark statistics). \pkg{UNSAT} specifications arise naturally during generation: a single constraint is rarely contradictory, whereas several combined constraints often are.

% =============================================================================
\subsection{Evaluation}
\label{sec:verification}

\paragraph{\bench{} evaluation.}
\label{sec:verifier}
The LogicVal evaluation comprises two steps: (1) extracting a scene graph $\widehat{S}(I)$ from the generated image $I$ and encoding it in ASP using predicates \texttt{object/1}, \texttt{hasProperty/3}, and \texttt{hasRelationship/3}; and (2) using \clingo{} to check whether the scene graph satisfies the constraints used to generate the image. We use Grounded-SAM (GSAM) \citep{ren2024grounded} for object detection and a zero-shot CLIP classifier \citep{radford2021clip} for attribute prediction. Appendix~\ref{app:metrics} details the scene graph extraction module.

%% file: tables/constraint_taxonomy.tex
\begin{table*}[t]
\centering
\tiny
\renewcommand{\arraystretch}{1.15}
\newcommand{\classrule}{\arrayrulecolor{black!25}\specialrule{0.3pt}{1pt}{1pt}\arrayrulecolor{black}}
\begin{tabularx}{\textwidth}{@{}l >{\raggedright\arraybackslash}p{0.085\textwidth} c >{\raggedright\arraybackslash}p{0.275\textwidth} >{\raggedright\arraybackslash\ttfamily}p{0.275\textwidth} >{\raggedright\arraybackslash}X@{}}
\toprule
\textbf{ID} & \textbf{Name} & \textbf{\#T} & \textbf{Abstract Form} & \textbf{\rmfamily Example template instance (ASP)} & \textbf{Natural language} \\
\midrule
\multicolumn{6}{>{\columncolor{basecolor}}l}{\emph{Object-level}}\\
\addlinespace[2pt]
C1 & Existential Object & 11 & $\exists x\,\Phi(x)$
   & :- \#count\{X: obj(X), hP(X,color,red)\} = 0.
   & A red object is in the scene. \\
\classrule
C2 & Universal Object & 9 & $\forall x\,\Phi(x)$
   & :- obj(X), hP(X,shape,sphere).
   & No object is sphere. \\
\classrule
C3 & Conditional Object & 10 & $\forall x\,(\Phi(x)\rightarrow\Psi(x))$
   & :- obj(X), hP(X,color,blue), hP(X,region,r3).
   & Every blue object is not in r3. \\
\midrule
\multicolumn{6}{>{\columncolor{basecolor}}l}{\emph{SubGraph-level}}\\
\addlinespace[2pt]
C4 & Existential & 4 & $\exists\bar{x}\,(\Phi(\bar{x})\wedge R(\bar{x}))$
   & :- \#count\{Z,X1,X2: \ldots, hP(Z,shape,sphere), hP(X1,region,r1), hP(X2,shape,cube), hR(Z,X1,right), hR(Z,X2,behind)\} = 0.
   & There exists a sphere object that is to the right of an object in r1, and also behind a cube object. \\
\classrule
C5 & Conditional & 6 & $\forall\bar{x}\,(\Phi(\bar{x})\wedge R(\bar{x})\rightarrow\Psi(\bar{x})\wedge R'(\bar{x}))$
   & :- obj(X1), obj(X2), X1!=X2, hR(X1,X2,right), not hP(X1,color,red).
   & Only objects that are red can be to the right of any object. \\
\classrule
C6 & Exist.-Universal & 4 & $\exists\bar{x}\,(\Phi(\bar{x})\wedge R(\bar{x})\wedge\forall\bar{y}\,(\Psi(\bar{y})\wedge R'(\bar{y})\rightarrow R(\bar{x},\bar{y})))$
   & cex(X) :- obj(X), obj(Y), X!=Y, hP(X,color,cyan), hP(Y,region,r1), not hR(X,Y,right).\newline
     ok :- obj(X), hP(X,color,cyan), not cex(X).\ \ \newline
     :- not ok.
   & There exists some cyan object such that it is to the right of every object in r1. \\
\classrule
C7 & Univ.-Existential & 6 & $\forall\bar{x}\,(\Phi(\bar{x})\wedge R(\bar{x})\rightarrow\exists\bar{y}\,(\Psi(\bar{y})\wedge R'(\bar{y})\wedge R(\bar{x},\bar{y})))$
   & :- obj(X), hP(X,shape,cube), \#count\{Y: obj(Y), X!=Y, hP(Y,shape,sphere), hR(Y,X,behind)\} = 0.
   & Every cube object has a sphere object behind it. \\
\midrule
\multicolumn{6}{>{\columncolor{basecolor}}l}{\emph{Cardinality-level}}\\
\addlinespace[2pt]
%C8 & Cardinality & 18 & $|\{x_1\mid\exists x_2,\ldots,x_t\,(\Phi(\bar{x})\wedge R(\bar{x}))\}|\;\theta\;k$
C8 & Cardinality & 18 & $\lvert\{x_1\mid\exists x_2,\ldots,x_t\,(\Phi(\bar{x})\wedge R(\bar{x}))\}\rvert\;\theta\;k$
   & :- \#count\{X: obj(X), hP(X,color,red)\} < 3.
   & There are at least three red objects. \\
\classrule
C9 & Aggregate Comp. & 3 & $\lvert\{x_1\mid\exists x_2,\ldots,x_t\,(\Phi(\bar{x})\wedge R(\bar{x}))\}\rvert$\newline $\theta\;\lvert\{y_1\mid\exists y_2,\ldots,y_m\,(\Psi(\bar{y})\wedge R'(\bar{y}))\}\rvert$
%C9 & Aggregate Comp. & 3 & $|\{x_1\mid\exists x_2,\ldots,x_t\,(\Phi(\bar{x})\wedge R(\bar{x}))\}|$\newline $\theta\;|\{y_1\mid\exists y_2,\ldots,y_m\,(\Psi(\bar{y})\wedge R'(\bar{y}))\}|$
   & l(N) :- N = \#count\{X: obj(X), hP(X,shape,sphere)\}.\newline
     r(M) :- M = \#count\{Y: obj(Y), hP(Y,color,red)\}.\newline
     :- l(N), r(M), N != M.
   & The count of sphere objects equals the count of red objects. \\
\bottomrule
\end{tabularx}
\caption{Primitive constraint classes with their abstract form (notation as in \S\ref{sec:taxonomy}), the number of templates in the class (\#T; Appendix~\ref{app:templates}), an example template from the class, and its natural language form. In the ASP column, \texttt{obj}, \texttt{hP} and \texttt{hR} abbreviate \texttt{object}, \texttt{hasProperty} and \texttt{hasRelationship}, auxiliary predicates are renamed, ``\ldots'' omits the object and distinctness atoms, and the anti-vacuity rules (and the corresponding part of the prompt).} 
\label{tab:taxonomy}
\end{table*}

%% file: sections/experimental_setup.tex
\section{Experimental Setup}
\label{sec:setup}

\subsection{Models}
We evaluate two proprietary T2I models, \gpt{} \citep{openai2026gptimage2} and \gemini{} \citep{google2025geminiimage}, and two open-weight models, \flux{} \citep{bfl2024flux} and \qwen{} \citep{wu2025qwen}. Each prompt consists of a fixed description of the CLEVR domain (the $2\times2$ regions, the allowed attribute values and the number of objects) followed by the verbalised constraint specification; the models also receive an instruction to generate an empty scene if the description is contradictory (prompts and generation settings in Appendix~\ref{app:prompts}).
We generate one image per prompt.

To help the open models produce CLEVR-style scenes with the requested vocabulary, we finetune them with LoRA \citep{hu2022lora} on CLEVR images \citep{johnson2017clevr} captioned with C1--C9 constraints that hold in them using the same natural-language templates as those used to generate the benchmark prompts (Appendix~\ref{app:training}).

\subsection{Human Evaluation}
\label{sec:annotation}
To obtain a reference verdict, two annotators shared the annotation of the scene graph $\Sgold(I)$ of all 3,887 generated images (\gpt{}: 1,018; \gemini{}: 833; \flux{}: 1,018; \qwen{}: 1,018); the counts differ because \gemini{} failed to generate an image for some prompts. Annotators used a dedicated web application to mark every object and its attributes (Appendix~\ref{app:annotation}).
The human verdict is $h(I)=1$ iff $\Sgold(I)\models C$, or, for an unsatisfiable $C$, iff $\Sgold(I)$ is empty; the \bench{} score $v(I)$ is defined in the same way on $\widehat{S}(I)$. The satisfaction rates in \S\ref{sec:results} are averages of $h(I)$.

\subsection{Automatic Evaluation Metrics}
%\paragraph{Reference-free alignment metrics.}
\label{sec:metrics-def}
We compare our \bench{} score with three widely used alignment metrics:
\noindent (i) \textbf{CLIPScore} \citep{hessel2021clipscore}: the raw cosine similarity between image and prompt embeddings. We use Long-CLIP \citep{zhang2024longclip} instead of standard CLIP, as it can handle our long prompts. The score ranges from $-1$ to $+1$, with higher values indicating greater semantic alignment between the generated image and the prompt.
\noindent (ii) \textbf{VLM-Judge} \cite{lee2024prometheusvision}: a vision-language model  rates prompt--image alignment from $1$ to $5$ using a fixed rubric, rescaled to $[0,1]$. Higher scores indicate greater perceived satisfaction of the prompt's constraints. We use GPT-4o \cite{hurst2024gpt} as the judge.
\noindent (iii) \textbf{Soft-TIFA} \citep{kamath2025geneval}: the prompt is decomposed into a set of sub-questions, each associated with an expected answer. A VQA model (GPT-4o) answers the sub-questions generated with each prompt, and each answer is scored by the probability mass assigned by the model to the correct answer. The scores of an image's sub-questions are combined by their arithmetic mean (AM), indicating atom-level correctness, or their geometric mean (GM), indicating prompt-level correctness; GM drops sharply if any single answer is wrong. The score ranges from $0$ to $1$, with higher values indicating greater alignment between the generated image and the prompt. Refer Appendix \ref{app:metrics} for more details about the implementation of the three metrics.
  
\subsection{Meta-Evaluation}\label{sec:metaeval}
We assess the alignment between automatic metrics 
and human judgments using the \textbf{point-biserial correlation} and the area under the receiver operating
characteristic curve \textbf{(ROC-AUC)}. Point-biserial correlation measures the strength and direction of the association between continuous metric scores and binary human judgments, while ROC-AUC measures how well a metric distinguishes between positive and negative human judgments across decision thresholds. Higher absolute point-biserial correlation indicates a stronger association, whereas a higher ROC-AUC indicates better discrimination with 0.5 corresponding to random performance and 1.0 to perfect discrimination.

We treat $h(I)$ as ground truth, calling images with $h(I)=1$ \emph{human-correct} ($I^{+}$) and the others \emph{human-incorrect} ($I^{-}$), and compare it with the score $s(I)$ of each metric (\S\ref{sec:metrics-def}); For every metric and model, we report
%(i)~the mean of $s$ over $I^{+}$ and over $I^{-}$;
(i)~the point-biserial correlation between $s$ and $h$; and
(ii)~the ROC-AUC between $s$ and $h$.
For the binary \bench{} score, we additionally report its \textbf{exact agreement} with $h(I)$, i.e.\ the fraction of images on which the two verdicts coincide.

%% file: sections/results.tex
\section{Results}
\label{sec:results}

%\subsection{How Well Do T2I Models Satisfy Constraints?}
%\subsubsection{1 Constraint at a time (1SAT)
\paragraph{Single-Constraint Specifications (\pkg{1SAT}).}
Table~\ref{tab:overall-main} reports, for each model and package, the percentage of images that satisfy all constraints of their prompt, as verified by human annotation. Appendix~\ref{app:qualitative} shows example images.
On \pkg{1SAT}, where each prompt specifies a single constraint in addition to the object-count condition, \gemini{} satisfies the prompt in 85\% of cases and \gpt{} in 58\%, whereas the LoRA-adapted open models succeed in fewer than 4\%.

\input{tables/overall_main.tex}
\input{tables/deformed_objects.tex}

For the open models, part of this gap is a rendering problem rather than a reasoning one.
Annotators flagged deformed or unrecognisable objects in 14\% of \flux{} and 51\% of \qwen{} images on \pkg{1SAT}, against at most 0.1\% for the proprietary models (Table~\ref{tab:deformed-objects}).
Even so, rendering alone does not explain the gap: \flux{} renders cleanly in 86\% of cases yet satisfies fewer than 4\% of the constraints.

\begin{figure}[h]
\centering
\includegraphics[width=0.85\columnwidth]{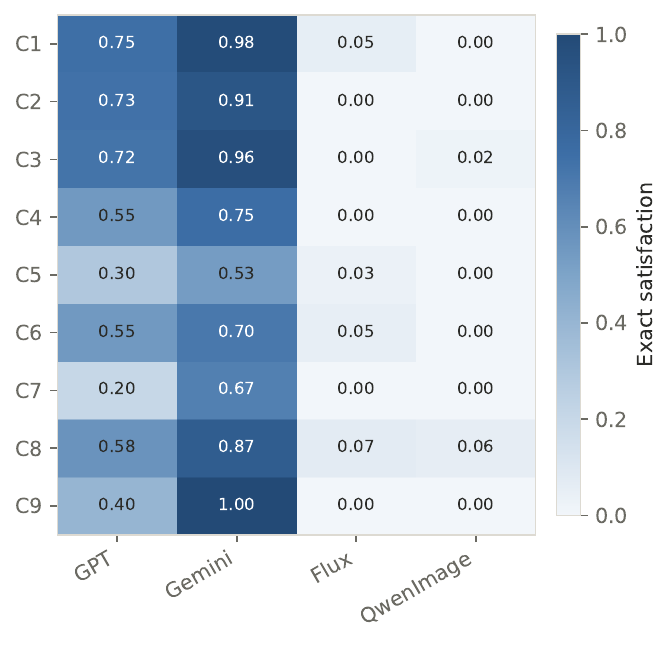}
%\caption{Share of generated images that satisfy their constraint, per constraint class (rows) and model (columns), on \pkg{1SAT} with short prompts, as verified by human annotation. C1--C3 constrain single objects, C4--C7 constrain relations between objects under existential and universal quantifiers, and C8--C9 constrain object counts. Darker cells mean more images satisfied the constraint.}
\caption{Success rate per constraint class (based on human judgement) on \pkg{1SAT}. C1--C3: single objects; C4--C7: relations under quantifiers; C8--C9: counts.}
\label{fig:class-heatmap}
\end{figure}

\paragraph{Effect of Logical Structure.}
Figure~\ref{fig:class-heatmap} breaks \pkg{1SAT} down by constraint class.
For both proprietary models, object-level classes (C1--C3) are the easiest: \gemini{} reaches 0.91--0.98 and \gpt{} 0.72--0.75.
The open models stay below 0.1 in every class.
Performance drops for subgraph-level constraints and is lowest on C5 (conditional subgraph: \gpt{} 0.30, \gemini{} 0.53) and C7 ($\forall\exists$: 0.20 and 0.67).
Both classes require the model to keep a relation consistent across \emph{every} object that matches a condition, which a single plausible arrangement does not guarantee.
The results per-template in Appendix~\ref{app:per-template} shows the same pattern at finer level: the hardest templates are those whose condition is triggered by a relation rather than an attribute.
On cardinality constraints the two models diverge: \gemini{} handles absolute counts (C8: 0.87) and count comparisons (C9: 1.00), whereas \gpt{} drops to 0.58 and 0.40.
This gap is consistent with the counting analysis in \S\ref{sec:counting}.

\subsection{Composing Multiple Constraints}
We investigate whether handling multiple constraints simultaneously poses a greater challenge. Our results show that conjoining constraints sharply lowers success. \gemini{} falls from 0.85 on one constraint to 0.48 on three (\pkg{3SAT}) and 0.20 on six (\pkg{6SAT}).
\gpt{} falls from 0.58 to 0.20 and to 0. Neither open model satisfies a single composed specification. 
These results suggest that satisfying multiple constraints requires explicit reasoning to guide object arrangement during generation.

\subsection{Recognising Unsatisfiable Specifications}
None of the tested T2I models reliably detects contradictory specifications. We evaluated their ability to recognise unsatisfiable constraints using 129 prompts with no valid solution. \gemini{} correctly identified 23.25\% of these specifications, abstaining by returning a blank image with a white background, as instructed. In contrast, the other models, including \gpt{}, almost never abstained (at most 1\%), instead generating scenes that necessarily violated at least one constraint.
The open models abstain on at most 2\% of specifications. %Recognising unsatisfiability requires reasoning about the specification before rendering it, and this capability is largely absent from current generators.

\subsection{Sensitivity to Prompt Wording}
\begin{figure}[h]
\centering
\includegraphics[width=\columnwidth]{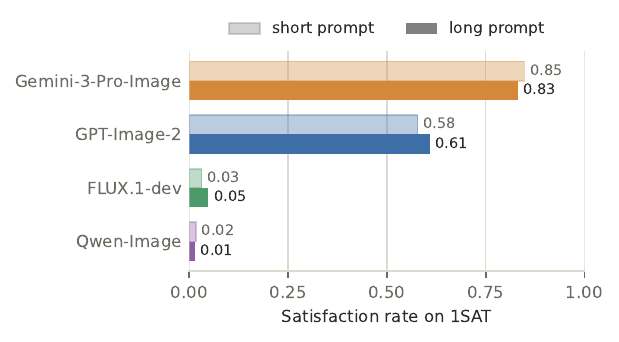}
\caption{Success rate on \pkg{1SAT}: short vs.\ long prompts.}
\label{fig:short-vs-long}
\end{figure}

On \pkg{1SAT}, rephrasing a prompt in greater detail - e.g., by explicitly specifying the quantifier scope, has little effect on performance (Figure~\ref{fig:short-vs-long}). While \gpt{} exhibits a modest improvement (0.58$\rightarrow$0.61), the changes for other models are negligible. Making constraints more explicit is no substitute for reasoning about them.

\subsection{Counting Errors}
\label{sec:counting}
Every prompt states the number of objects in the scene, followed by a logical constraint drawn from C1 to C9 that must be satisfied by the objects. For example, a prompt might state: ``\textit{There are 5 objects in the scene. At least 2 are red}''. A model may generate an image containing at least two red objects while failing to produce five objects.

Table~\ref{tab:overall-main} and Figure~\ref{fig:class-heatmap} require both the logical constraint and the object count to be satisfied. Here, we relax the count requirement to test whether models satisfy the logical constraint even when they generate the wrong number of objects. This reveals cases in which models satisfy the requested relations and attributes but not the object count.

\begin{figure}[t]
\centering
\includegraphics[width=\columnwidth]{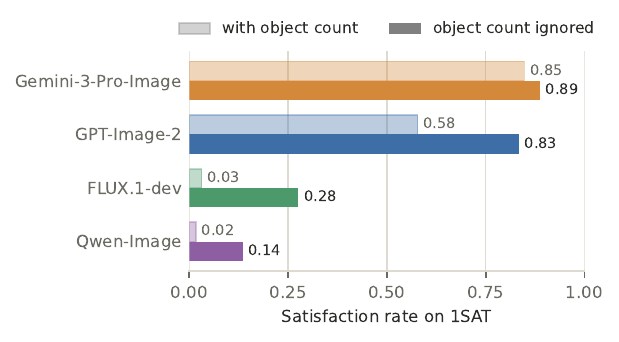}
\caption{Satisfaction rate on \pkg{1SAT}  with and without the object-count requirement.}
\label{fig:count-vs-no-count}
\end{figure}

Figure~\ref{fig:count-vs-no-count} shows improvements under this relaxed evaluation. \gpt{}'s success rate rises from 0.58 to 0.83, while \gemini{}'s increases from 0.85 to 0.89. Object counting also limits \gpt{}'s performance on the cardinality classes (C8 and C9), whereas \gemini{} is already strong on them (C8: 0.87, C9: 1.00). The gains are more pronounced for \flux{} (0.03$\rightarrow$0.28) and \qwen{} (0.02$\rightarrow$0.14).
%For \gpt{}, failure to match the requested object count accounts for 26 percentage points of the 42-percentage-point gap between perfect performance and its original success rate, even when the specified relations and attributes are correct.

%% file: tables/overall_main.tex
\begin{table}[h]
\centering
\scriptsize
\begin{tabular}{lcccc}
\toprule
 & GPT & Gemini & Flux & QwenImage \\
\midrule
1SAT & 0.577 & \textbf{0.848} & 0.031 & 0.017 \\
3SAT & 0.200 & \textbf{0.480} & 0.000 & 0.000 \\
6SAT & 0.000 & \textbf{0.200} & 0.000 & 0.000 \\
%1UNSAT & 0.000 & \textbf{0.220} & 0.000 & 0.020 \\
%3UNSAT & 0.000 & \textbf{0.200} & 0.000 & 0.000 \\
%6UNSAT & 0.000 & \textbf{0.246} & 0.000 & 0.000 \\
\bottomrule
\end{tabular}
%\caption{Share of generated images that satisfy their constraints, per model (columns) and benchmark package (rows), as verified by human annotation of each image. On SAT packages an image counts as correct only if its scene graph satisfies every constraint, including the requested number of objects; The leading digit is the number of constraints combined in one prompt (1, 3 or 6), Best model per row in bold.}
\caption{Success rate per model and package (based on human judgement).}
\label{tab:overall-main}
\end{table}

%% file: tables/deformed_objects.tex
\begin{table}[h]
\centering
\scriptsize
\begin{tabular}{lcccc}
\toprule
 & GPT & Gemini & Flux & QwenImage \\
\midrule
1SAT & 0.001 & 0.000 & 0.138 & 0.507 \\
3SAT & 0.000 & 0.000 & 0.040 & 0.120 \\
6SAT & 0.000 & 0.000 & 0.040 & 0.040 \\
\bottomrule
\end{tabular}
%\caption{Share of generated images that were flagged during human annotation as containing at least one deformed or unrecognisable object.}
\caption{Rate of images with at least one deformed object (based on human judgement).}
\label{tab:deformed-objects}
\end{table}

%% file: sections/metric_reliability.tex
%\section{How Reliable Are Automatic Metrics?}
\begin{table}[t]
\centering
\scriptsize
\setlength{\tabcolsep}{4pt}
\resizebox{\columnwidth}{!}{%
\begin{tabular}{lcccc}
\toprule
 Metric& GPT & Gemini & Flux & QwenImage \\
\midrule
ClipScore      & -0.081 & 0.158 & 0.046 & 0.009 \\
VLM-Judge      &  0.127 & 0.152 & 0.217 & 0.160 \\
Soft-TIFA (AM) &  0.247 & 0.128 & 0.276 & 0.223 \\
Soft-TIFA (GM) &  0.289 & 0.216 & 0.246 & 0.206 \\
LogicVal       &  \textbf{0.970} & \textbf{0.700} & \textbf{0.712} & \textbf{0.522 }\\
\bottomrule
\end{tabular}%
}
\caption{Point-biserial correlation with human scores.}
\label{tab:point-biserial}
\end{table}

\section{Reliability of Automatic Metrics}
\label{sec:metrics}

\begin{figure}[t]
\centering
\includegraphics[width=\columnwidth]{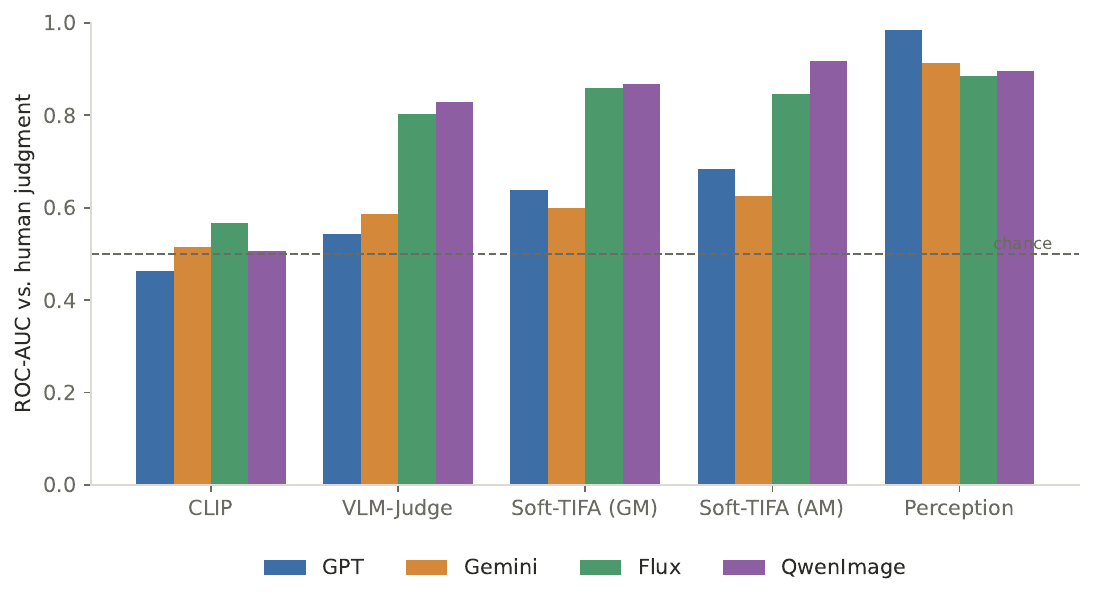}
\caption{ROC-AUC of each metric against the human verdict on \pkg{1SAT}; the dashed line marks chance (0.5).}
\label{fig:roc-1sat}
\end{figure}

Human scene-graph annotation is costly, motivating an automatic proxy. We evaluate metrics by their alignment with human verdicts.
Table \ref{tab:point-biserial} shows point-biserial correlation and Figure~\ref{fig:roc-1sat} reports ROC-AUC on \pkg{1SAT} (710 images per model).
\textbf{CLIPScore is at chance level} for every model (0.46--0.57).
It scores satisfying and violating images almost identically (0.22 on average for both).
VLM-Judge and Soft-TIFA depend on the T2I model.
They differentiate correct from incorrect images reasonably well for \flux{} and \qwen{} (AUC 0.80--0.92).
For \gpt{} and \gemini{} they are close to chance (0.54--0.68).
The open models' failures are blatant: missing or deformed objects, or the wrong scene altogether.
The proprietary models' failures are subtle: for example a relation being reversed.
The VLM-based metrics catch the first kind and miss the second, so they are least informative exactly where the models are strongest.
\textbf{LogicVal} reaches an AUC of 0.88--0.98 for all four models.

\begin{table}[h]
\centering
\footnotesize
\begin{tabular}{lcccc}
\toprule
 & GPT & Gemini & Flux & QwenImage \\
\midrule
%1SAT & 96.338 &	90.7042 &	97.6056	& 96.6197 \\
1SAT & 96.34 &	90.70 &	97.61	& 96.62 \\
3SAT & 96	& 88	& 100 &	100	\\
6SAT & 100	& 92& 100	& 100 \\
%1UNSAT & 0.000 & \textbf{0.220} & 0.000 & 0.020 \\
%3UNSAT & 0.000 & \textbf{0.200} & 0.000 & 0.000 \\
%6UNSAT & 0.000 & \textbf{0.246} & 0.000 & 0.000 \\
\bottomrule
\end{tabular}
%\caption{\bench{} vs.\ human verdicts (\%).}
\caption{Percentage of images both \bench{} and human accept or reject.}
\label{tab:exact}
\end{table}

Since both LogicVal and human scores are binary, we show the percentage of exact agreement between the scores in Table \ref{tab:exact}.
The errors in LogicVal judgement are mostly false positives - in other words, it accepts scenes that humans reject.

\begin{table}[h]
\centering
\footnotesize
\begin{tabular}{lcccc}
\toprule
 & GPT & Gemini & Flux & QwenImage \\
\midrule
1SAT & \textbf{0.744} & 0.323 & 0.154 & 0.141 \\
3SAT & \textbf{0.600} & 0.560 & 0.040 & 0.200 \\
6SAT & 0.480 & \textbf{0.680} & 0.160 & 0.240 \\
\bottomrule
\end{tabular}
%\caption{Accuracy of the automatic perception, defined as the share of images whose automatically extracted object list exactly matches the human annotation.}
\caption{Scene Graph Extraction Accuracy: percentage of predicted graphs matching human annotations.}
\label{tab:perception-accuracy}
\end{table}

Exact recovery of the full object list (count, colour, shape, material and region) also degrades when the image generated is complex due to factors like object occlusion. Table~\ref{tab:perception-accuracy} shows the accuracy in extracting the scene graph. However, only the attributes mentioned in a constraint affect its final verdict, so the verdict can be right even when the entire object list is not.
This highlights the importance of improving object detection and attribute prediction to ensure reliable perception and, ultimately, more accurate LogicVal scores.

%% file: sections/conclusion.tex
\section{Conclusion}
\label{sec:conclusion}

We formulate text-to-image generation under explicit logical requirements as a constraint-satisfaction problem and introduce \bench{}, a solver-labelled benchmark for evaluating constraint satisfaction through scene-graph verification. Our results show that current generators struggle with exact counts, universally quantified relations, and composed constraints, and rarely recognise contradictions. Widely used alignment metrics fail to capture these shortcomings for the strongest models, whereas \bench{} provides a more effective, though still imperfect, means of assessing constraint satisfaction. In future work, we aim to extend LogicVal to a probabilistic formulation that accounts for uncertainty in object attribute prediction. Future work will also explore generation pipelines that integrate probabilistic LogicVal scores for solver-guided feedback.

%% file: sections/limitations.tex
\newpage
\section{Limitations}
\paragraph{Synthetic Domain.} \bench{} is instantiated in the CLEVR domain, enabling controlled evaluation of logical constraints but limiting coverage of real-world concepts and scenes. Future work will extend the framework to domains such as healthcare, architecture, and virtual try-on.

\paragraph{Human Annotation Errors.} Human judgement is susceptible to perceptual errors in colour, material, and spatial relations. Human judgments may, for example, confuse cyan with blue or misidentify materials and object positions. Resource constraints limited annotation to two annotators (two of the authors), who labelled non-overlapping subsets of images, precluding inter-annotator agreement analysis. 

%\paragraph{Limitations of LogicVal as Automatic Verifier.} 
\paragraph{Limitations of LogicVal score.} 
LogicVal score's performance degrades on complex scenes involving occlusion or deformed objects. For example, some generated shapes, particularly from open models, appeared to combine features of cylinders and cubes, leading the detector to assign one of these labels to shapes that fit neither category. Reliable evaluation also depends on improvements in both domain-specific detection models, particularly when extending LogicVal to domains such as healthcare, and open-vocabulary object detection and attribute classification.

\paragraph{Model Configurations.} For cost reasons, proprietary models were evaluated at lower-cost settings (\gpt{} at a low-quality tier and \gemini{} at 1K resolution), with one image generated per prompt. Higher-quality settings or best-of-n sampling could change the absolute results.

%% file: sections/ethics.tex
\section{Ethical considerations}

\bench{} contains only synthetic geometric scenes and templated prompts; it involves no personal data and no depictions of people.
Proprietary models were accessed through their public APIs in accordance with their terms of use.
%\todo{annotators: who they were, how they were recruited and compensated, and that they were informed about the purpose of the annotation}

The 3,887 generated images were annotated by two of the authors, who were fully aware of the purpose of the annotation; no crowdworkers or external annotators were involved, and the images contain only synthetic geometric objects.

The code is available at \repolink{} under the MIT license; the benchmark, the annotations and the generated images will be released under CC~BY~4.0 with the camera-ready version.

%% file: sections/appendix/background_program.tex
\section{Background ASP Program}
\label{app:background}

Every program solved by \clingo{} consists of a domain specification, the background program, and the instantiated constraint templates (Table~\ref{tab:taxonomy}).

\paragraph{Domain}
The domain identifies the type of objects, attributes and possible values for each attribute.
The CLEVR domain is represented in ASP as:

\begin{codebox}[basicstyle=\ttfamily\scriptsize, escapeinside={(*}{*)}]

property(color,gray). property(color,red).
property(color,blue). property(color,green).
property(color,brown). property(color,cyan).
property(color,purple). property(color,yellow).
property(shape,cube). property(shape,sphere).
property(shape,cylinder).
property(material,rubber).
property(material,metal).
property(region,r0). property(region,r1).
property(region,r2). property(region,r3).

1 { hasProperty(X,color,V) :
      property(color,V) } 1 :- object(X).
1 { hasProperty(X,shape,V) :
      property(shape,V) } 1 :- object(X).
1 { hasProperty(X,material,V) :
      property(material,V) } 1 :- object(X).
1 { hasProperty(X,region,V) :
      property(region,V) } 1 :- object(X).
\end{codebox}

The spatial relations in CLEVR are defined as follows:
\begin{codebox}[basicstyle=\ttfamily\scriptsize, escapeinside={(*}{*)}]
% Region layout (2x2 grid)
% r0=top-left | r1=top-right
% r2=bot-left | r3=bot-right
right_R(r1,r0). right_R(r3,r0).
right_R(r1,r2). right_R(r3,r2).
left_R(R1,R2) :- right_R(R2,R1).
front_R(r2,r0). front_R(r3,r0).
front_R(r2,r1). front_R(r3,r1).
behind_R(R1,R2) :- front_R(R2,R1).

% Relations derived from the regions
hasRelationship(X,Y,left) :-
    object(X), object(Y), X != Y,
    hasProperty(X,region,R1),
    hasProperty(Y,region,R2), left_R(R1,R2).
hasRelationship(X,Y,right) :-
    object(X), object(Y), X != Y,
    hasProperty(X,region,R1),
    hasProperty(Y,region,R2), right_R(R1,R2).
hasRelationship(X,Y,front) :-
    object(X), object(Y), X != Y,
    hasProperty(X,region,R1),
    hasProperty(Y,region,R2), front_R(R1,R2).
hasRelationship(X,Y,behind) :-
    object(X), object(Y), X != Y,
    hasProperty(X,region,R1),
    hasProperty(Y,region,R2),
    behind_R(R1,R2).

% Every pair is ordered on both axes
hasRelationship(X,Y,left);
  hasRelationship(X,Y,right) :-
    object(X), object(Y), X != Y,
    hasProperty(X,region,R1),
    hasProperty(Y,region,R2).
hasRelationship(X,Y,front);
  hasRelationship(X,Y,behind) :-
    object(X), object(Y), X != Y,
    hasProperty(X,region,R1),
    hasProperty(Y,region,R2).

% Anti-symmetry
:- hasRelationship(X1,X2,R),
   hasRelationship(X2,X1,R), X1 != X2.

% Inverse
hasRelationship(X1,X2,left) :-
    hasRelationship(X2,X1,right).
hasRelationship(X1,X2,right) :-
    hasRelationship(X2,X1,left).
hasRelationship(X1,X2,front) :-
    hasRelationship(X2,X1,behind).
hasRelationship(X1,X2,behind) :-
    hasRelationship(X2,X1,front).

% Transitivity
hasRelationship(X1,X3,left) :-
    hasRelationship(X1,X2,left),
    hasRelationship(X2,X3,left),
    X1 != X2, X2 != X3.
hasRelationship(X1,X3,behind) :-
    hasRelationship(X1,X2,behind),
    hasRelationship(X2,X3,behind),
    X1 != X2, X2 != X3.
\end{codebox}

 The above definition allows deriving the spatial relations between objects based on the regions they are placed. For example, since region \textit{r0} is left of region \textit{r1}, an object placed in the former region is also left of an object placed in the latter region. Every pair of distinct objects is ordered along both axes, 
following anti-symmetry, inverse and transitivity axioms.

%% file: sections/appendix/constraint_templates.tex
\section{ASP Constraint Templates}
\label{app:templates}

The constraint templates are represented in ASP. An ASP template has placeholders for properties (\texttt{P'}), values (\texttt{V'}), directions (\texttt{D'}) and counts (\texttt{N'}), that are filled at random; comparisons such as \texttt{V1'!=V2'} exclude degenerate instances.
%In the names, \texttt{$k$prop} gives the number of properties, \texttt{neg}/\texttt{mix} mark negated/mixed literals, \texttt{A}/\texttt{C} the antecedent/consequent, and \texttt{atleast}/\texttt{atmost}/\texttt{exact} the comparison.
Rules marked \texttt{anti-vacuity} require at least one object that triggers the condition of a universal constraint; without them, a scene could be vacuously true, for instance, scene could satisfy ``every blue sphere is metal'' simply by containing no blue sphere. Following is the list of templates.

\begin{codebox}[basicstyle=\ttfamily\scriptsize, escapeinside={(*}{*)}]
(*\textbf{C1\_1prop}*)
:- #count { X : object(X),
    hasProperty(X, P1', V1')} = 0.

(*\textbf{C1\_1prop\_2val\_neg}*)
:- #count {X: object(X),
    not hasProperty(X, P1', V1'),
    not hasProperty(X, P1', V2'), V1'!=V2'} = 0.

(*\textbf{C1\_1prop\_3val\_neg}*)
:- #count {X: object(X),
    not hasProperty(X, P1', V1'),
    not hasProperty(X, P1', V2'),
    not hasProperty(X, P1', V3'), V1'!=V2',
    V1'!=V3', V2'!=V3'} = 0.

(*\textbf{C1\_1prop\_neg}*)
:- #count {X: object(X),
    not hasProperty(X, P1', V1')} = 0.

(*\textbf{C1\_2prop}*)
:- #count { X : object(X),
    hasProperty(X, P1', V1'),
    hasProperty(X, P2', V2'), P1'!=P2'} = 0.

(*\textbf{C1\_2prop\_mix\_neg}*)
:- #count {X: object(X), hasProperty(X, P1', V1'),
    not hasProperty(X, P2', V2'), P1'!=P2'} = 0.

(*\textbf{C1\_2prop\_neg}*)
:- #count {X: object(X),
    not hasProperty(X, P1', V1'),
    not hasProperty(X, P2', V2'), P1'!=P2'} = 0.

(*\textbf{C1\_3prop}*)
:- #count { X : object(X),
    hasProperty(X, P1', V1'),
    hasProperty(X, P2', V2'),
    hasProperty(X, P3', V3'), P1'!=P2', P1'!=P3',
    P2'!=P3'} = 0.

(*\textbf{C1\_3prop\_val\_mix\_neg}*)
:- #count {X: object(X), hasProperty(X, P1', V1'),
    hasProperty(X, P2', V2'),
    not hasProperty(X, P3', V3'),
    not hasProperty(X, P3', V4'), P1'!=P2',
    P1'!=P3', P2'!=P3', V3'!=V4'} = 0.

(*\textbf{C1\_4prop}*)
:- #count { X : object(X),
    hasProperty(X, P1', V1'),
    hasProperty(X, P2', V2'),
    hasProperty(X, P3', V3'),
    hasProperty(X, P4', V4'), P1'!=P2', P1'!=P3',
    P1'!=P4', P2'!=P3', P2'!=P4', P3'!=P4'} = 0.

(*\textbf{C1\_4prop\_mix\_neg}*)
:- #count {X: object(X), hasProperty(X, P1', V1'),
    hasProperty(X, P2', V2'),
    not hasProperty(X, P3', V3'),
    not hasProperty(X, P4', V4'), P1'!=P2',
    P1'!=P3', P1'!=P4', P2'!=P3', P2'!=P4',
    P3'!=P4'} = 0.

(*\textbf{C2\_1prop}*)
:- object(X), not hasProperty(X, P1', V1').

(*\textbf{C2\_1prop\_2val\_neg}*)
:- object(X), V1'!=V2',
    1 {hasProperty(X, P1', V1');
    hasProperty(X, P1', V2')}.

(*\textbf{C2\_1prop\_neg}*)
:- object(X), hasProperty(X, P1', V1').

(*\textbf{C2\_2prop}*)
:- object(X), P1'!= P2',
    #count{1 : hasProperty(X, P1', V1');
    2 : hasProperty(X, P2', V2')} != 2.

(*\textbf{C2\_2prop\_mix\_neg}*)
:- #count{X: object(X), hasProperty(X, P1', V1'),
    not hasProperty(X, P2', V2'), P1'!=P2'}>0.

(*\textbf{C2\_2prop\_neg}*)
:- #count{X: object(X), hasProperty(X, P1', V1'),
    hasProperty(X, P2', V2'), P1'!=P2'}>0.

(*\textbf{C2\_3prop}*)
:- object(X), P1'!=P2', P2'!=P3', P1'!=P3',
    #count{1 : hasProperty(X, P1', V1');
    2 : hasProperty(X, P2', V2');
    3 : hasProperty(X, P3', V3')} != 3.

(*\textbf{C2\_4prop}*)
:- object(X), P1'!=P2', P1'!=P3', P1'!=P4',
    P2'!=P3', P2'!=P4', P3'!=P4',
    #count{1 : hasProperty(X, P1', V1');
    2 : hasProperty(X, P2', V2');
    3 : hasProperty(X, P3', V3');
    4 : hasProperty(X, P4', V4')} != 4.

(*\textbf{C2\_4prop\_neg}*)
:- #count{X: object(X), hasProperty(X, P1', V1'),
    hasProperty(X, P2', V2'),
    not hasProperty(X, P3', V3'),
    not hasProperty(X, P4', V4'), P1'!=P2',
    P1'!=P3', P1'!=P4', P2'!=P3', P2'!=P4',
    P3'!=P4'}>0.

(*\textbf{C3\_1propA\_1propC}*)
:- object(X), hasProperty(X, P1', V1'),
    not hasProperty(X, P2', V2').
% anti-vacuity
:- #count {X: object(X),
    hasProperty(X, P1', V1')} = 0.

(*\textbf{C3\_1propA\_1prop\_neg}*)
:- object(X), hasProperty(X, P1', V1'),
    hasProperty(X, P2', V2').
% anti-vacuity
:- #count {X: object(X),
    hasProperty(X, P1', V1')} = 0.

(*\textbf{C3\_1propA\_2propC}*)
:- object(X), hasProperty(X, P1', V1'), P1'!=P2',
    P1'!=P3', P2'!=P3',
    #count{ 1 : hasProperty(X, P2', V2');
    2 : hasProperty(X, P3', V3') } != 2.
% anti-vacuity
:- #count {X: object(X),
    hasProperty(X, P1', V1')} = 0.

(*\textbf{C3\_1propA\_3propC}*)
:- object(X), hasProperty(X, P1', V1'), P1'!=P2',
    P1'!=P3', P1'!=P4', P2'!=P3', P2'!=P4',
    P3'!=P4',
    #count{ 1 : hasProperty(X, P2', V2');
    2 : hasProperty(X, P3', V3');
    3 : hasProperty(X, P4', V4')} != 3.
% anti-vacuity
:- #count {X: object(X),
    hasProperty(X, P1', V1')} = 0.

(*\textbf{C3\_1propA\_neg\_1propC}*)
:- object(X), not hasProperty(X, P1', V1'),
    not hasProperty(X, P2', V2').
% anti-vacuity
:- #count {X: object(X),
    not hasProperty(X, P1', V1')} = 0.

(*\textbf{C3\_2propA\_1propC}*)
:- object(X), hasProperty(X, P1', V1'),
    hasProperty(X, P2', V2'),
    not hasProperty(X, P3', V3'), P1'!=P2',
    P1'!=P3', P2'!=P3'.
% anti-vacuity
:- #count {X: object(X), hasProperty(X, P1', V1'),
    hasProperty(X, P2', V2')} = 0.

(*\textbf{C3\_2propA\_2propC}*)
:- object(X), hasProperty(X, P1', V1'),
    hasProperty(X, P2', V2'), P1'!=P2', P1'!=P3',
    P2'!=P3', P1'!=P4', P2'!=P4', P3'!=P4',
    #count{ 1 : hasProperty(X, P3', V3');
    2 : hasProperty(X, P4', V4') } != 2.
% anti-vacuity
:- #count {X: object(X), hasProperty(X, P1', V1'),
    hasProperty(X, P2', V2')} = 0.

(*\textbf{C3\_2propA\_neg\_2propC}*)
:- object(X), hasProperty(X, P1', V1'),
    not hasProperty(X, P2', V2'), P1'!=P2',
    P1'!=P3', P2'!=P3', P1'!=P4', P2'!=P4',
    P3'!=P4',
    #count{ 1 : hasProperty(X, P3', V3');
    2 : hasProperty(X, P4', V4') } != 2.
% anti-vacuity
:- #count {X: object(X), hasProperty(X, P1', V1'),
    not hasProperty(X, P2', V2')} = 0.

(*\textbf{C3\_3propA\_1propC}*)
:- object(X), hasProperty(X, P1', V1'),
    hasProperty(X, P2', V2'),
    hasProperty(X, P3', V3'),
    not hasProperty(X, P4', V4'), P1'!=P2',
    P1'!=P3', P2'!=P3', P1'!=P4', P2'!=P4',
    P3'!=P4'.
% anti-vacuity
:- #count {X: object(X), hasProperty(X, P1', V1'),
    hasProperty(X, P2', V2'),
    hasProperty(X, P3', V3')} = 0.

(*\textbf{C3\_3propA\_1propC\_neg}*)
:- object(X), hasProperty(X, P1', V1'),
    hasProperty(X, P2', V2'),
    hasProperty(X, P3', V3'),
    hasProperty(X, P4', V4'), P1'!=P2', P1'!=P3',
    P2'!=P3', P1'!=P4', P2'!=P4', P3'!=P4'.
% anti-vacuity
:- #count {X: object(X), hasProperty(X, P1', V1'),
    hasProperty(X, P2', V2'),
    hasProperty(X, P3', V3')} = 0.

(*\textbf{C4\_1prop\_2hop}*)
:- #count { X1, X2, X3 : object(X1), object(X2),
    object(X3), X1 != X2, X2 != X3, X1 != X3,
    hasProperty(X1, P1', V1'),
    hasProperty(X2, P2', V2'),
    hasProperty(X3, P3', V3'),
    hasRelationship(X1, X2, D1'),
    hasRelationship(X2, X3, D2'), V1'!=V2',
    V1'!=V3', V2'!=V3' } = 0.

(*\textbf{C4\_1prop\_2hop\_mix}*)
:- #count { X1, X2, X3 : object(X1), object(X2),
    object(X3), X1 != X2, X2 != X3, X1 != X3,
    hasProperty(X1, P1', V1'),
    hasProperty(X2, P2', V2'),
    not hasProperty(X3, P3', V3'),
    hasRelationship(X1, X2, D1'),
    hasRelationship(X2, X3, D2'), V1'!=V2'} = 0.

(*\textbf{C4\_1prop\_shared}*)
:- #count {Z, X1, X2 : object(Z), object(X1),
    object(X2), Z != X1, Z != X2, X1 != X2,
    hasProperty(X1, P1', V1'),
    hasProperty(X2, P2', V2'),
    hasProperty(Z, P3', V3'),
    hasRelationship(Z, X1, D1'),
    hasRelationship(Z, X2, D2'), V1'!=V2',
    V1'!=V3', V2'!=V3' } = 0.

(*\textbf{C4\_1prop\_shared\_mix}*)
:- #count {Z, X1, X2 : object(Z), object(X1),
    object(X2), Z != X1, Z != X2, X1 != X2,
    hasProperty(X1, P1', V1'),
    not hasProperty(X2, P2', V2'),
    hasProperty(Z, P3', V3'),
    hasRelationship(Z, X1, D1'),
    hasRelationship(Z, X2, D2'), V1'!=V3' } = 0.

(*\textbf{C5\_pair\_propA\_relC}*)
:- object(X1), object(X2), X1 != X2,
    hasProperty(X1, P1', V1'),
    hasProperty(X2, P2', V2'), V1'!=V2',
    not hasRelationship(X1, X2, D1').
% anti-vacuity
:- #count {X1, X2: object(X1), object(X2),
    X1!= X2, hasProperty(X1, P1', V1'),
    hasProperty(X2, P2', V2'), V1'!=V2'} = 0.

(*\textbf{C5\_pair\_propRelA\_RelC}*)
:- object(X1), object(X2), X1 != X2,
    hasProperty(X1, P1', V1'),
    hasProperty(X2, P2', V2'), V1'!=V2',
    hasRelationship(X1, X2, D1'),
    not hasRelationship(X1, X2, D2').
% anti-vacuity
:- #count {X1, X2: object(X1), object(X2),
    X1 != X2, hasProperty(X1, P1', V1'),
    hasProperty(X2, P2', V2'),
    hasRelationship(X1, X2, D1'), V1'!=V2'} = 0.

(*\textbf{C5\_pair\_propRelA\_propC}*)
:- object(X1), object(X2), X1 != X2,
    hasProperty(X1, P1', V1'),
    hasProperty(X2, P2', V2'),
    hasRelationship(X1, X2, D1'),
    not hasProperty(X2, P3', V3'), V1'!=V2',
    P3'!=P2'.
% anti-vacuity
:- #count {X1, X2: object(X1), object(X2),
    X1 != X2, hasProperty(X1, P1', V1'),
    hasProperty(X2, P2', V2'),
    hasRelationship(X1, X2, D1'), V1'!=V2'} = 0.

(*\textbf{C5\_pair\_relA\_propC}*)
:- object(X1), object(X2), X1 != X2,
    hasRelationship(X1, X2, D1'),
    not hasProperty(X1, P1', V1').
% anti-vacuity
:- #count {X1, X2: object(X1), object(X2),
    X1 != X2, hasRelationship(X1, X2, D1')} = 0.

(*\textbf{C5\_triple\_propA\_RelC}*)
:- object(X1), object(X2), object(X3), X1!=X2,
    X2!=X3, X1!=X3, hasProperty(X1, P1', V1'),
    hasProperty(X2, P2', V2'),
    hasProperty(X3, P3', V3'), V1'!=V2', V2'!=V3',
    V1'!=V3',
    #count{ 1 : hasRelationship(X1, X2, D1');
    2 : hasRelationship(X2, X3, D2') } != 2.
% anti-vacuity
:- #count {X1, X2, X3: object(X1), object(X2),
    object(X3), X1!=X2, X1!=X3, X2!=X3,
    hasProperty(X1, P1', V1'),
    hasProperty(X2, P2', V2'),
    hasProperty(X3, P3', V3'), V1'!=V2', V2'!=V3',
    V1'!=V3'} = 0.

(*\textbf{C5\_triple\_propRelA\_relC}*)
:- object(X1), object(X2), object(X3), X1!= X2,
    X2!= X3, X1!= X3, hasProperty(X1, P1', V1'),
    hasProperty(X2, P2', V2'),
    hasProperty(X3, P3', V3'),
    hasRelationship(X1, X2, D1'),
    not hasRelationship(X2, X3, D2'), V1' != V2',
    V2' != V3', V1' != V3'.
% anti-vacuity
:- #count {X1, X2, X3: object(X1), object(X2),
    object(X3), X1!=X2, X1!=X3, X2!=X3,
    hasProperty(X1, P1', V1'),
    hasProperty(X2, P2', V2'),
    hasProperty(X3, P3', V3'),
    hasRelationship(X1, X2, D1'), V1'!=V2',
    V2'!=V3', V1'!=V3'} = 0.

(*\textbf{C6\_1prop}*)
c6_1_prop_counterexample(X) :- object(X),
    object(Y), X!=Y, hasProperty(X, P1', V1'),
    hasProperty(Y, P2', V2'),
    not hasRelationship(X, Y, D1'), V1'!=V2'.
c6_1_prop_satisfied :- object(X),
    hasProperty(X, P1', V1'),
    not c6_1_prop_counterexample(X).
:- not c6_1_prop_satisfied.
% anti-vacuity
:- #count{Y: object(Y),
    hasProperty(Y, P2', V2')}=0.

(*\textbf{C6\_1prop\_neg}*)
c6_forbid_1_prop_counterexample(X) :- object(X),
    object(Y), X!=Y, hasProperty(X, P1', V1'),
    not hasProperty(Y, P2', V2'),
    not hasRelationship(X, Y, D1').
c6_forbid_1_prop_satisfied :- object(X),
    hasProperty(X, P1', V1'),
    not c6_forbid_1_prop_counterexample(X).
:- not c6_forbid_1_prop_satisfied.
% anti-vacuity
:- #count{Y: object(Y),
    not hasProperty(Y, P2', V2')}=0.

(*\textbf{C6\_2prop}*)
c6_2_prop_counterexample(X) :- object(X),
    object(Y), X!=Y, hasProperty(X, P1', V1'),
    hasProperty(X, P2', V2'),
    hasProperty(Y, P3', V3'),
    hasProperty(Y, P4', V4'),
    not hasRelationship(X, Y, D1'), P1'!=P2',
    P3'!=P4', V1'!=V3', V1'!=V4', V2'!=V3',
    V2'!=V4'.
c6_2_prop_satisfied :- object(X),
    hasProperty(X, P1', V1'),
    hasProperty(X, P2', V2'), P1'!=P2',
    not c6_2_prop_counterexample(X).
:- not c6_2_prop_satisfied.
% anti-vacuity
:- #count{Y: object(Y), hasProperty(Y, P3', V3'),
    hasProperty(Y, P4', V4'), P3'!=P4'}=0.

(*\textbf{C6\_2prop\_neg}*)
c6_forbid_2_prop_counterexample(X) :- object(X),
    object(Y), X!=Y, hasProperty(X, P1', V1'),
    hasProperty(X, P2', V2'),
    hasProperty(Y, P3', V3'),
    not hasProperty(Y, P4', V4'),
    not hasRelationship(X, Y, D1'), P1'!=P2',
    P3'!=P4', V1'!=V3', V2'!=V3'.
c6_forbid_2_prop_satisfied :- object(X),
    hasProperty(X, P1', V1'),
    hasProperty(X, P2', V2'), P1'!=P2',
    not c6_forbid_2_prop_counterexample(X).
:- not c6_forbid_2_prop_satisfied.
% anti-vacuity
:- #count{Y: object(Y), hasProperty(Y, P3', V3'),
    not hasProperty(Y, P4', V4'), P3'!=P4'}=0.

(*\textbf{C7\_1prop\_exact}*)
:- object(X), hasProperty(X, P1', V1'),
    #count { Y : object(Y), X != Y,
    hasProperty(Y, P2', V2'),
    hasRelationship(Y, X, D1'), V1'!=V2'} != N'.
% anti-vacuity
:- #count {X: object(X),
    hasProperty(X, P1', V1')} = 0.

(*\textbf{C7\_1prop\_exact\_neg}*)
:- object(X), hasProperty(X, P1', V1'),
    #count { Y : object(Y), X != Y,
    not hasProperty(Y, P2', V2'),
    hasRelationship(Y, X, D1') } != N'.
% anti-vacuity
:- #count {X: object(X),
    hasProperty(X, P1', V1')} = 0.

(*\textbf{C7\_1prop\_propRel}*)
:- object(X), hasProperty(X, P1', V1'),
    #count { Y : object(Y), X != Y,
    hasProperty(Y, P2', V2'), V2'!=V1',
    hasRelationship(Y, X, D1') } = 0.
% anti-vacuity
:- #count {X: object(X),
    hasProperty(X, P1', V1')} = 0.

(*\textbf{C7\_1prop\_propRel\_neg}*)
:- object(X), not hasProperty(X, P1', V1'),
    #count { Y : object(Y), X != Y,
    hasProperty(Y, P2', V2'),
    hasRelationship(Y, X, D1') } = 0.
% anti-vacuity
:- #count {X: object(X),
    not hasProperty(X, P1', V1')} = 0.

(*\textbf{C7\_propRel\_propRel}*)
:- object(X), object(Z), hasProperty(X, P1', V1'),
    hasProperty(Z, P2', V2'), V1'!=V2',
    hasRelationship(X, Z, D1'), X!=Z,
    #count { Y : object(Y), X != Y, Y!=Z,
    hasProperty(Y, P3', V3'), V1'!=V3', V2'!=V3',
    hasRelationship(Y, X, D2') } = 0.
% anti-vacuity
:- #count {X, Z: object(X), object(Z),
    hasProperty(X, P1', V1'),
    hasProperty(Z, P2', V2'), V1'!=V2',
    hasRelationship(X, Z, D1'), X!=Z} = 0.

(*\textbf{C7\_propRel\_propRel\_neg}*)
:- object(X), object(Z), hasProperty(X, P1', V1'),
    hasProperty(Z, P2', V2'), V1'!=V2',
    hasRelationship(X, Z, D1'), X!=Z,
    #count { Y : object(Y), X != Y, Y!=Z,
    not hasProperty(Y, P3', V3'),
    hasRelationship(Y, X, D2') } = 0.
% anti-vacuity
:- #count {X, Z: object(X), object(Z),
    hasProperty(X, P1', V1'),
    hasProperty(Z, P2', V2'), V1'!=V2',
    hasRelationship(X, Z, D1'), X!=Z} = 0.

(*\textbf{C8\_1prop\_atleast}*)
:- #count{X: object(X),
    hasProperty(X, P1', V1')} < N'.

(*\textbf{C8\_1prop\_atleast\_neg}*)
:- #count{X: object(X),
    not hasProperty(X, P1', V1')} < N'.

(*\textbf{C8\_1prop\_atleast\_relational}*)
:- #count{X: object(X), hasProperty(X, P1', V1'),
    object(Y), X != Y, hasProperty(Y, P2', V2'),
    hasRelationship(X, Y, D1'), V1'!=V2'} < N'.

(*\textbf{C8\_1prop\_atleast\_relational\_neg}*)
:- #count{X: object(X), hasProperty(X, P1', V1'),
    object(Y), X != Y, hasProperty(Y, P2', V2'),
    not hasRelationship(X, Y, D1'),
    V1'!=V2'} < N'.

(*\textbf{C8\_1prop\_atmost}*)
:- #count{X: object(X),
    hasProperty(X, P1', V1')} > N'.

(*\textbf{C8\_1prop\_atmost\_neg}*)
:- #count{X: object(X),
    not hasProperty(X, P1', V1')} > N'.

(*\textbf{C8\_1prop\_atmost\_relational}*)
:- #count{X: object(X), hasProperty(X, P1', V1'),
    object(Y), X != Y, hasProperty(Y, P2', V2'),
    hasRelationship(X, Y, D1'), V1'!=V2'} > N'.

(*\textbf{C8\_1prop\_atmost\_relational\_neg}*)
:- #count{X: object(X), hasProperty(X, P1', V1'),
    object(Y), X != Y, hasProperty(Y, P2', V2'),
    not hasRelationship(X, Y, D1'),
    V1'!=V2'} > N'.

(*\textbf{C8\_1prop\_exact}*)
:- #count{X: object(X),
    hasProperty(X, P1', V1')} != N'.

(*\textbf{C8\_1prop\_exact\_neg}*)
:- #count{X: object(X),
    not hasProperty(X, P1', V1')} != N'.

(*\textbf{C8\_1prop\_exact\_relational}*)
:- #count{X: object(X), hasProperty(X, P1', V1'),
    object(Y), X != Y, hasProperty(Y, P2', V2'),
    hasRelationship(X, Y, D1'), V1'!=V2'} != N'.

(*\textbf{C8\_1prop\_exact\_relational\_neg}*)
:- #count{X: object(X), hasProperty(X, P1', V1'),
    object(Y), X != Y, hasProperty(Y, P2', V2'),
    not hasRelationship(X, Y, D1'),
    V1'!=V2'} != N'.

(*\textbf{C8\_2prop\_atleast}*)
:- #count{X: object(X), hasProperty(X, P1', V1'),
    hasProperty(X, P2', V2'), P1'!=P2'} < N'.

(*\textbf{C8\_2prop\_atleast\_neg}*)
:- #count{X: object(X), hasProperty(X, P1', V1'),
    not hasProperty(X, P2', V2'), P1'!=P2'} < N'.

(*\textbf{C8\_2prop\_atmost}*)
:- #count{X: object(X), hasProperty(X, P1', V1'),
    hasProperty(X, P2', V2'), P1'!=P2'} > N'.

(*\textbf{C8\_2prop\_atmost\_neg}*)
:- #count{X: object(X), hasProperty(X, P1', V1'),
    not hasProperty(X, P2', V2'), P1'!=P2'} > N'.

(*\textbf{C8\_2prop\_exact}*)
:- #count{X: object(X), hasProperty(X, P1', V1'),
    hasProperty(X, P2', V2'), P1'!=P2'} != N'.

(*\textbf{C8\_2prop\_exact\_neg}*)
:- #count{X: object(X), hasProperty(X, P1', V1'),
    not hasProperty(X, P2', V2'), P1'!=P2'} != N'.

(*\textbf{C9\_1prop}*)
c9_1prop_left(N, V1') :-
    N = #count { X : object(X),
    hasProperty(X, P1', V1') }.
c9_1prop_right(N, V2') :-
    N = #count { Y : object(Y),
    hasProperty(Y, P2', V2') }.
:- c9_1prop_left(N, V1'), c9_1prop_right(M, V2'),
    N!= M, V1'!=V2'.
% anti-vacuity
:- #count {X : object(X),
    hasProperty(X, P1', V1')} = 0.

(*\textbf{C9\_2prop}*)
c9_2prop_left(N, V1') :-
    N = #count { X : object(X),
    hasProperty(X, P1', V1'),
    hasProperty(X, P2', V2'), P1'!=P2'}.
c9_2prop_right(N, V3') :-
    N = #count { Y : object(Y),
    hasProperty(Y, P3', V3'),
    hasProperty(Y, P4', V4'), P4'!=P3'}.
:- c9_2prop_left(N, V1'), c9_2prop_right(M, V3'),
    N!= M, V1'!=V3'.
% anti-vacuity
:- #count {X : object(X),
    hasProperty(X, P1', V1'),
    hasProperty(X, P2', V2'), P1'!=P2'} = 0.

(*\textbf{C9\_2prop\_mix}*)
c9_2prop_mix_left(N, V2') :-
    N = #count { X : object(X),
    not hasProperty(X, P1', V1'),
    hasProperty(X, P2', V2'), P1'!=P2'}.
c9_2prop_mix_right(N, V4') :-
    N = #count { Y : object(Y),
    not hasProperty(Y, P3', V3'),
    hasProperty(Y, P4', V4'), P4'!=P3'}.
:- c9_2prop_mix_left(N, V2'),
    c9_2prop_mix_right(M, V4'), N!= M, V2'!=V4'.
% anti-vacuity
:- #count {X : object(X),
    not hasProperty(X, P1', V1'),
    hasProperty(X, P2', V2'), P1'!=P2'} = 0.
\end{codebox}

%% file: sections/appendix/prompts.tex
\section{Prompt for Image Generation}
\label{app:prompts}

\paragraph{Prompt template.}
Every prompt consists of a CLEVR domain description, the number of objects, the constraint specification, and an instruction to generate an empty white scene if the description is contradictory.
Only the proprietary models are also asked for a white background (in italics); the open models are not, since they are fine-tuned on CLEVR images, whose background is not white:
\begin{quote}\small
Generate a CLEVR style scene \emph{with a white background,} that is divided into 4 regions arranged in a 2x2 grid: r0 (top-left); r1 (top-right); r2 (bottom-left); r3 (bottom-right). Each object has a color (one of: gray, red, blue, green, brown, purple, cyan, yellow), a shape (one of: cube, cylinder, sphere), a material (one of: rubber, metal), and is placed in one of the 4 regions. The scene must additionally satisfy the following constraints: The scene is with \textbf{$\langle$ Add the number of objects here $\rangle$} objects. \textbf{$\langle$Add the constraint to be satisfied here$\rangle$} If the scene description is invalid or contains a contradiction, generate an empty scene with white background.
\end{quote}

For multiple-constraint specifications, the different constraints are concatenated as a numbered list (``Constraint 1: \dots{} Constraint 2: \dots''). Table~\ref{tab:prompt-examples} shows some examples of constraint specifications that are placed at the \textbf{$\langle$Add the constraint to be satisfied here$\rangle$} slot in the above prompt format.

\paragraph{Generation settings.}
\gpt{} is used at the \emph{low} quality tier and $1024\times1024$, and \gemini{} at 1K resolution.
\flux{} (28 steps, guidance 3.5) and \qwen{} (50 steps, true-CFG scale 4.0) generate at $1024\times688$ with their selected LoRA weights (Appendix~\ref{app:training}).

\begin{table*}[t]
\centering
\tiny
\begin{tabularx}{\textwidth}{@{}lX@{}}
\toprule
\multicolumn{2}{@{}l}{\textbf{C3 (template \texttt{1propA\_1propC}), $n=4$}}\\
ASP & \texttt{:- object(X), hasProperty(X,color,green), not hasProperty(X,material,metal).}\newline\texttt{:- \#count\{X: object(X), hasProperty(X,color,green)\} = 0.}\\
Short & At least one object is green, and every green object is also metal.\\
Long & The scene must contain at least one object that is green. Whenever an object is green, it must simultaneously be metal.\\
\midrule
\multicolumn{2}{@{}l}{\textbf{C7 (template \texttt{1prop\_exact}), $n=6$}}\\
ASP & \texttt{:- object(X), hasProperty(X,region,r1), \#count\{Y: object(Y), X!=Y, hasProperty(Y,material,rubber), hasRelationship(Y,X,behind)\} != 3.}\newline\texttt{:- \#count\{X: object(X), hasProperty(X,region,r1)\} = 0.}\\
Short & There must be at least one object that is in r1. Every such object has exactly three rubber objects that are behind it.\\
Long & The scene must contain at least one object that is in r1. Every object that is in r1 is required to have exactly three witnesses: A witness is a rubber object standing behind it.\\
\midrule
\multicolumn{2}{@{}l}{\textbf{C9 (template \texttt{1prop}), $n=6$}}\\
ASP & \texttt{left(N) :- N = \#count\{X: object(X), hasProperty(X,shape,cylinder)\}.}\newline\texttt{right(N) :- N = \#count\{Y: object(Y), hasProperty(Y,color,cyan)\}.}\newline\texttt{:- left(N), right(M), N != M.}\quad\texttt{:- \#count\{X: object(X), hasProperty(X,shape,cylinder)\} = 0.}\\
Short & There must be at least one cylinder object. The count of cylinder objects equals the count of cyan objects.\\
Long & The scene must contain at least one cylinder object. The scene enforces a balance: the cardinality of objects that are cylinder must exactly match the cardinality of objects that are cyan.\\
\bottomrule
\end{tabularx}
\caption{Three \pkg{1SAT} specifications, each shown with its instantiated ASP rules, its two natural-language forms - a short and a long version. The second ASP rule of each example is the anti-vacuity condition.}
\label{tab:prompt-examples} 
\end{table*}

%% file: sections/appendix/benchmark_statistics.tex
\section{Benchmark Statistics}
\label{app:bench-stats}

Table~\ref{tab:bench-stats} gives the number of prompts in each package (1, 3 and 6 SAT/UNSAT). 
\pkg{1SAT} is constructed by generating 5 instantiations of each of the 71 templates ($71*5$ $=$$355$ prompts).Both \pkg{3SAT} and \pkg{6SAT} have 25 prompts each. Each of the 25 instances are constructed by combining 3 or 6 template instantiations (each template from a different class). 
The \pkg{UNSAT}, 3SAT and 6SAT result from random sampling and are not balanced across classes.
Scenes contain 3--9 objects.
For each \pkg{SAT} specification, one answer set found by \clingo{} is stored as a hidden reference scene; it is never shown to the generators.

\begin{table}[h]
\centering
\scriptsize
\begin{tabular}{@{}lrrr@{}}
\toprule
Constraints per spec ($m$) & \pkg{SAT} & \pkg{UNSAT} & Total \\
\midrule
1 & 355 & 25 & 380 \\
3 & 25  & 10 & 35 \\
6 & 25  & 69 & 94 \\
\midrule
Total & 405 & 104 & 509 \\
\addlinespace[2pt]
Templates covered (\pkg{1SAT}) & 71 & 14 & \\
\bottomrule
\end{tabular}

\vspace{8pt}
\setlength{\tabcolsep}{2.4pt}
\begin{tabular}{@{}lrrrrrrrrrr@{}}
\toprule
Class & C1 & C2 & C3 & C4 & C5 & C6 & C7 & C8 & C9 & Total \\
\midrule
Templates & 11 & 9 & 10 & 4 & 6 & 4 & 6 & 18 & 3 & 71 \\
\pkg{1SAT} prompts & 55 & 45 & 50 & 20 & 30 & 20 & 30 & 90 & 15 & 355 \\
\bottomrule
\end{tabular}
\caption{Number of prompts per package (top) and per constraint class in \pkg{1SAT} (bottom).}
\label{tab:bench-stats}
\end{table}

%% file: sections/appendix/verifier_and_metrics.tex
\section{Evaluation Metric Details}
\label{app:metrics}

\paragraph{\bench{} Score.}
\bench{} evaluation involves a perception stage that extracts a scene graph from the generated image. The resulting scene graph is then passed to the CLINGO solver alongside the constraints used to generate the image to determine whether they are satisfied. In this section, we describe the perception pipeline used to construct the scene graph.

\noindent \textbf{Object Detection and Bounding Boxes}: We use Grounding DINO \citealp{ren2024grounded} for bounding-box (BB) detection, with a confidence threshold of 0.15. To remove duplicate detections, we apply a post-processing step that suppresses a lower-confidence detection when either: its intersection over union (IoU) with a higher-confidence detection exceeds a specified threshold or it is mostly contained within a higher-confidence detection.

\noindent \textbf{Attribute Classification}
We use the CLIP model \citep{radford2021clip} (openai/clip-vit-large-patch14) for zero-shot attribute classification. Multiple views or crops of each detected object are classified using different prompt templates to improve robustness.
\begin{itemize}
    \item \textbf{Prompt-template aggregation}: For each image view, we classify the crop using multiple prompt templates and average the resulting logits across templates.
    \item \textbf{Image-view aggregation}: We average the aggregated logits across multiple views of the same object. The views depend on the attribute being predicted: shape - tight and medium crops, with a grey background, colour - tight, medium, and wide crops, with a grey background, material - medium crops, with the original background, to preserve relevant contextual cues such as shadows and reflections.
\end{itemize}
We apply softmax to the averaged logits and select the class with the highest probability using argmax.

\noindent \textbf{Spatial Grounding}: An object's region is the quadrant containing its bounding-box centre. The spatial relations between objects are also inferred based on the coordinates of the bounding box center. 

\begin{table*}[t]
\centering
\scriptsize
\renewcommand{\arraystretch}{1.2}
\begin{tabularx}{\textwidth}{
    @{}
    >{\raggedright\arraybackslash}p{0.17\textwidth}
    >{\raggedright\arraybackslash}X
    >{\raggedright\arraybackslash}p{0.20\textwidth}
    @{}
}
\toprule
\textbf{Category} & \textbf{Sub-question} & \textbf{Expected answer} \\
\midrule

\multicolumn{3}{@{}l}{\textbf{Common sub-questions}}\\
\midrule

Common (SAT)
& Is there any shape other than cube, sphere, or cylinder?
& No \\

Common (SAT)
& Are all objects among the following colors: gray, red, blue, green, brown, purple, cyan, or yellow?
& Yes \\

Common (SAT)
& True or False: All objects are either metal or rubber.
& True \\

Common (SAT)
& Is the image an empty scene with a white background?
& No \\

Common (SAT)
& How many objects are in the scene?
& \texttt{number\_of\_objects} (from the prompt) \\

Common (UNSAT)
& Is the image an empty scene with a white background?
& Yes \\

\midrule
\multicolumn{3}{@{}l}{\textbf{Prompt-specific sub-questions}}\\
\midrule

\multicolumn{3}{@{}p{\textwidth}}{Prompt: The scene must contain at least one object that is green. Whenever an object is green, it must also be metal.
}\\
\midrule

Sub-QA
& How many objects are green?
& $c>0$ \\

Sub-QA
& How many objects are both green and metal?
& $=c$ \\

Sub-QA
& How many objects are green but not metal?
& $=0$ \\

Sub-QA
& True or False: There is at least one green object that is not metal.
& False \\

\midrule
\multicolumn{3}{@{}p{\textwidth}}{Prompt:
There must be at least one object that is in r1. Every such object has exactly three rubber objects that are behind it.
}\\
\midrule

Sub-QA
& How many objects that are in r1 are in the image?
& $c>0$ \\

Sub-QA
& True or False: If there is an object that is in r1, there is a rubber object standing behind it.
& True \\

Sub-QA
& If there is an object that is in r1, how many rubber objects are standing behind it?
& $=3$ \\

\bottomrule
\end{tabularx}
\caption{Common and prompt-specific sub-questions used in Soft-TIFA, along with their expected answers. Expected answers may be template-based, as in $c>0$, where the value of c is obtained from GPT-4o. Subsequent references to $c$, such as in $=c$, are instantiated using the value assigned to $c$ based on GPT-4o’s answer to the first sub-question.}
\label{tab:soft-tifa1}
\end{table*}

\paragraph{CLIPScore.}
We use LongCLIP (\texttt{LongCLIP-GmP-ViT-L-14}; \citealp{zhang2024longclip}), which supports text inputs of up to 248 tokens, compared with the 77-token limit of standard CLIP. Prompts in the 1SAT package fall within this limit, whereas those in the 3SAT and 6SAT packages exceed it. Tokens beyond the model's input limit are truncated during scoring, potentially resulting in scores that do not reflect the full prompt.
We report the raw cosine similarity between the image and text CLIP embeddings, without the usual rescaling.

\paragraph{VLM-Judge.}
The judge, GPT-4o,  receives the image and the following instruction:

\begin{quote}\small
You are judging whether a generated image matches a text description. Description: ``$\langle$prompt$\rangle$''. Rate how well the image satisfies all the logical constraints in the description on a scale of 1 (not at all) to 5 (perfect match). Respond in exactly this format: SCORE: $\langle$1--5$\rangle$ RATIONALE: $\langle$one sentence$\rangle$
\end{quote}

The score returned is then rescaled to $[0,1]$. 

\paragraph{Soft-TIFA.} \cite{kamath2025geneval} 
evaluates the correctness of a generated image by decomposing the original prompt into sub-questions and assessing the image based on the answers to these questions.  We implement Soft-TIFA using GPT-4o as the visual question answering (VQA) backend.

Since our prompts are generated by instantiating logical constraint template, each constraint template is also associated with a template of sub-questions and expected answers. These templates are instantiated alongside the constraint template instantiation, yielding the corresponding sub-questions and expected answers for each prompt. See Table \ref{tab:soft-tifa1} for examples. Rather than generating a complete textual response to each sub-question, the VQA model predicts a single token conditioned on the image, question, and set of candidate answers. The candidate answer set depends on the question type. For binary questions, the candidates are {yes,no} or {true,false}. For property-value questions, the candidates correspond to the possible values of the queried property. For cardinality questions, the candidate set consists of possible object counts, with its size determined by $\max(N,10)$, where $N$ is the number of objects specified in the prompt.

 We retrieve the top-20 next-token log-probabilities and convert them into probabilities. For each candidate answer, we account for possible token variants and aggregate their probability mass to obtain a soft score for the expected answer.

For questions that ask whether the number of objects is greater than zero, the score is the total probability assigned to positive counts, i.e., $1-P(0)$, assuming the count distribution covers all possible outcomes. 

\emph{Count ranges.} For count-comparison questions, the score is the sum of the probabilities assigned to all counts that satisfy the specified comparison. For example, for the question ``How many objects are green?'' with the expected answer being a condition like $c>0$, a model assigning $P(0)=0.10$, $P(1)=0.60$, and $P(2)=0.30$ receives a score of $P(1)+P(2)=0.90$.

\emph{Linked counts.} Some sub-questions refer to a count introduced by an earlier sub-question. For example, `\textit{How many objects are red?}' with the expected answer being a condition $c>0$ introduces the count $c$, while the second sub-question for the prompt `\textit{How many objects are both red and cubes?}' with the expected answer $=c$ refers to that count. Because the value of $c$ is not known in advance, we instantiate it using the model's most likely count predicted for the first question. For example, if the most likely answer is 2, the second question is scored using the probability assigned to the count 2.
The original Soft-TIFA used fixed expected answers (and not conditions like  $c>0$) and does not explicitly resolve dependencies between sub-questions.

%% file: sections/appendix/lora_training.tex
\section{LoRA Adaptation of the Open Models}
\label{app:training}

\paragraph{Captions.}
For every CLEVR training image \citep{johnson2017clevr}, the objects, attributes and pairwise relations between objects are read from the original scene graph annotations; no pixels are inspected.
For each class C1--C9, a grounder searches this scene graph for a combination of facts that makes one of the class's patterns true. For example, if the scene graph indicates there is a red cylinder, the template, 

\begin{figure*}[t]
\centering
\includegraphics[width=0.92\textwidth]{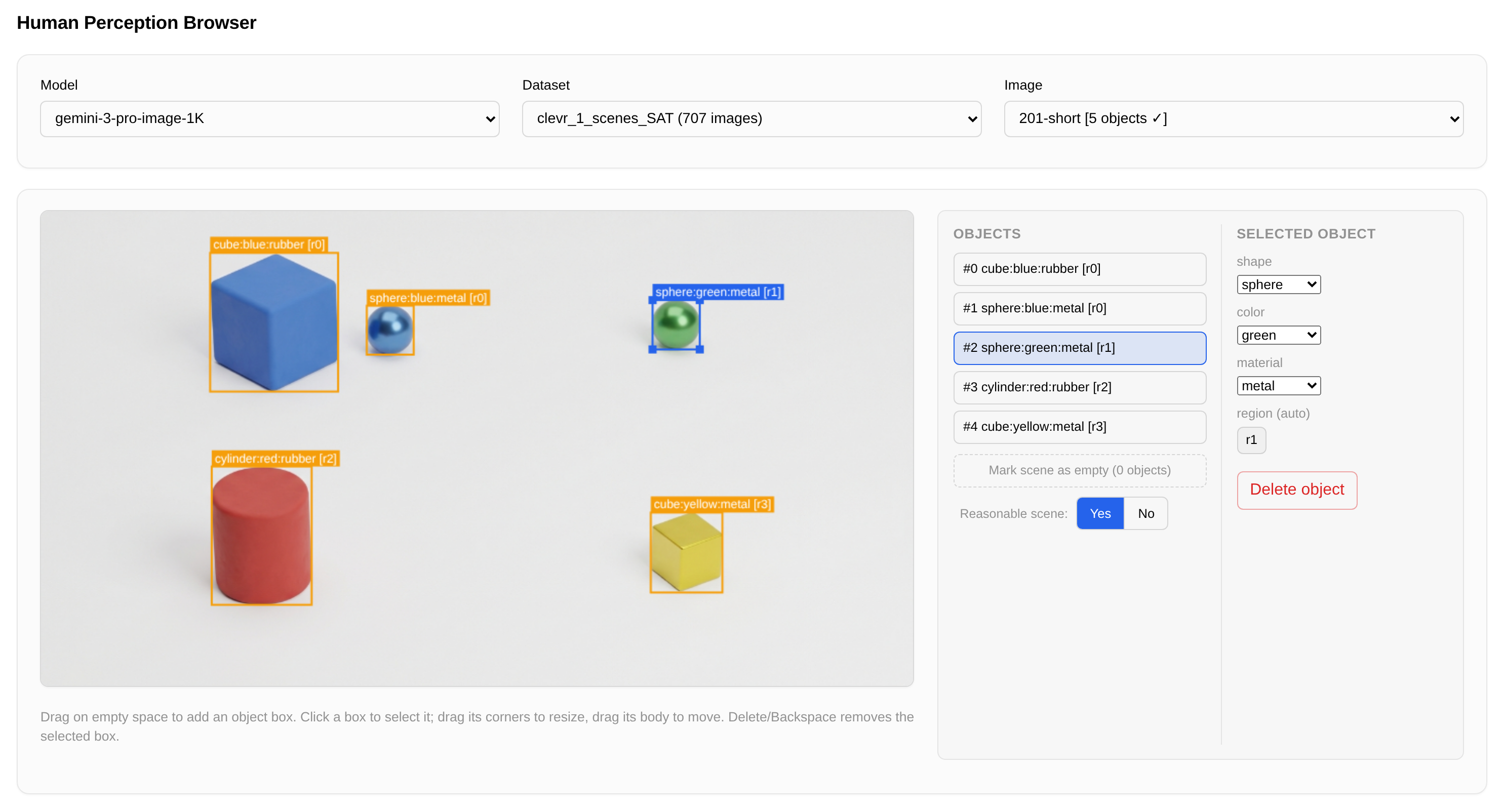}
\caption{The annotation tool, showing the Gemini-3-Pro-Image image of Figure~\ref{fig:overview}. The object list (centre) and the property editor (right) set each object's shape, colour and material, and its region is derived from the box. The annotator also marks whether the image is a reasonable scene, or marks it as empty.}
\label{fig:annotation-tool}
\end{figure*}

\begin{codebox}[basicstyle=\ttfamily\scriptsize, escapeinside={(*}{*)}]
(*\textbf{C1\_1prop}*)
:- #count { X : object(X), hasProperty(X, P1', V1')} = 0.
\end{codebox}

\noindent finds a match with the instantiation \texttt{(P1}$'$: \texttt{color}, \texttt{V1}$'$: \texttt{red)}. The instantiated template is then translated into it's natural language form with the benchmark's verbaliser. Every image is paired with a prompt that is generated in a similar way. Every prompt begins with a sentence that specifies the number of objects in the image followed by one or more constraints, e.g.\ \emph{``There are 6 objects in the scene. There must be at least one red object in the scene.''}. The prompts are balanced across all constraint classes, with each constraint class instance occurring an equal number of times in the training set.

\paragraph{Batches and selection.}
The \texttt{(prompt,image)} pairs are split into batches of about 2{,}000.
Each batch targets the same number of instances for every one of the 61 (class, template) pairs that occur in the training data, and no image is reused in a batch.
Training continues from one batch to the next, initialised from the previous checkpoint.
After every 50 steps, the loss is computed on a fixed, class-balanced held-out set of 99 images (11 per class) that never appear in any batch.
We select the checkpoint with the lowest held-out loss: \flux{} after batch~1 (held-out loss 0.280) and \qwen{} after batch~3 (0.048), both at 1{,}000 steps.

\paragraph{Hyperparameters.}
LoRA rank 16, $\alpha=16$, learning rate $10^{-4}$, batch size 1, resolution $1024\times688$, seed 42.
The training objective is the models' native flow-matching loss.
Training ran on a single NVIDIA RTX PRO 6000 GPU (32 vCPUs, 188\,GB RAM); one batch of about 2{,}000 samples took about four hours.

%% file: sections/appendix/human_annotation.tex
\section{Human Annotation}
\label{app:annotation}

\paragraph{Annotation tool.}
The annotation was carried out by two annotators, working in a browser-based web application that shows one generated image at a time.
The web application was partially developed using Claude Code \citep{soldani2026claude}.
The application allows the annotator to select a T2I model, a benchmark package and an image generated by the T2I model for one of the prompts in the selected package from drop-down menus. The annotator can then edit the annotation present for the image, and every edit is saved automatically. The tool runs as a password-protected web service, which allowed both annotators to work remotely on the same files.

Figure~\ref{fig:annotation-tool} shows the annotation view.

\paragraph{What is annotated.}
\begin{itemize}[leftmargin=*,nosep]
  \item \textbf{Objects.} The annotator draws one bounding box per object and selects its shape, colour and material from the closed CLEVR vocabulary (size is not annotated; see \S\ref{sec:benchmark}).
  \item \textbf{Spatial information.} Regions and spatial relations are not entered by hand. It can be inferred from the annotated bounding box.
  \item \textbf{Scene flags.} The annotator marks whether the image has rendering errors and flags images containing deformed or unrecognisable objects; these flags yield the deformation rates in Table~\ref{tab:deformed-objects}. An image without any object is marked as an empty scene, which is the correct output for an unsatisfiable specification.
\end{itemize}

\paragraph{From annotation to verdict.}
For each image, the annotation is stored as a scene graph in the same format as the automatic perception output.
It is converted into ASP facts (\texttt{object/1}, \texttt{hasProperty/3}, \texttt{hasRelationship/3}) and solved together with the instantiated specification by \clingo{}; the image is judged correct if the solver finds the specification satisfied (or, for an unsatisfiable specification, if the scene is empty).

Annotators did not see the prompt or the constraint while annotating.
Each image was annotated by one of the two annotators and took about one minute.

%% file: sections/appendix/additional_results.tex
\input{figures/qualitative-example/qualitative_examples}

\begin{figure*}[h]
\centering
\includegraphics[width=\textwidth]{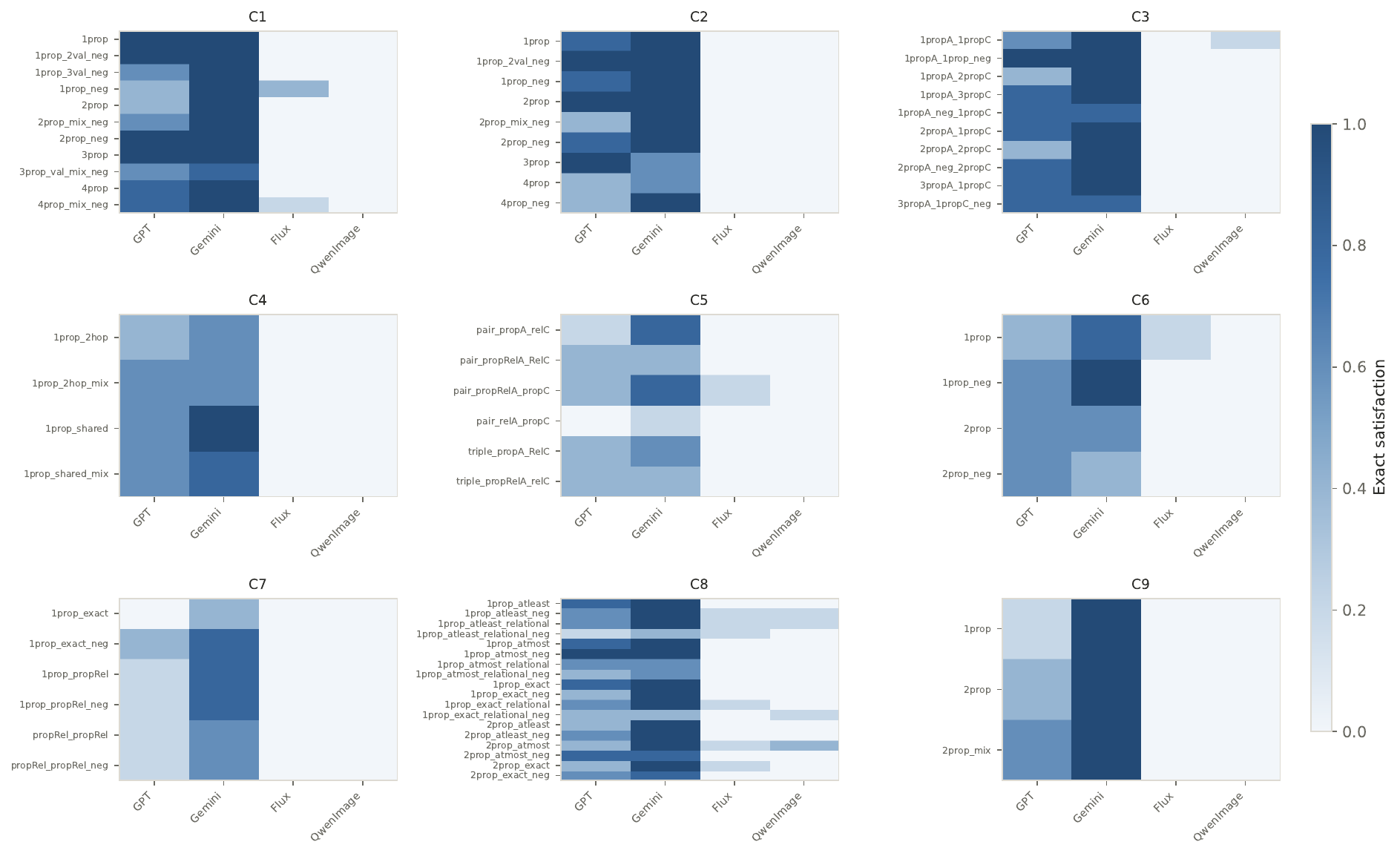}
\caption{Satisfaction rate per template on \pkg{1SAT} (short prompts), one panel per class; each cell is computed from five images.}
\label{fig:family-heatmap}
\end{figure*}
\section{Additional Results}
\label{app:additional}

\subsection{Qualitative Examples}
\label{app:qualitative}
Figure~\ref{fig:qualitative} shows the images that the four T2I models generated for the same prompt, with the human verdict under each image and, for failures, the reason.
The rows follow the Results section: a single attribute constraint and a relational constraint (\S\ref{sec:results}), a constraint that is satisfied but with the wrong number of objects (\S\ref{sec:counting}), a composed specification (\pkg{3SAT}), and an unsatisfiable specification, for which only \gemini{} returns the expected empty scene.

\subsection{Per-Template Results}
\label{app:per-template}

Figure~\ref{fig:family-heatmap} breaks the \pkg{1SAT} results down by template.

\gemini{} satisfies most object-level (C1--C3) and plain counting (C8, C9) templates perfectly; its failures concentrate on templates that tie a requirement to a \emph{relation}.
The hardest template for both proprietary models is C5 \texttt{pair\_relA\_propC} (\gpt{}: 0.0, \gemini{}: 0.2), which requires every object that stands in a given relation to another one to have a property, so the condition is triggered by the layout of the scene rather than by an attribute.
Statements like `\textit{For all x there exists exactly n y objects such that ... }' is similarly hard (C7 \texttt{1prop\_exact}: 0.0 and 0.4).
Within C8, adding a relational context reduces the performance of \gemini{} from 1.0 on most plain counts to 0.4--0.6 on four of its six relational templates, including all three negated ones.

\gpt{} shows a sharper split.
It is perfect on nine templates, all object-level except one counting template (C8 \texttt{1prop\_atmost\_neg}), and lies between 0.4 and 0.8 on most other object-level and counting templates.
On the subgraph classes C4--C7 it never exceeds 0.6, and five of its six C7 templates (\texttt{1prop\_exact} and the four \texttt{propRel} templates) score 0.2 or lower, as does C5 \texttt{pair\_relA\_propC} (0.0).
Unlike \gemini{}, it also struggles with count comparison: on C9 \texttt{1prop}, which requires two sets to be equally large, it reaches 0.2 against \gemini{}'s 1.0.

Negation does not make templates uniformly harder: for \gpt{} on C1, the negated two-attribute template (\texttt{2prop\_neg}, 1.0) is easier than its positive counterpart (\texttt{2prop}, 0.4), while the single-attribute one shows the opposite (\texttt{1prop} 1.0 vs.\ \texttt{1prop\_neg} 0.4).
Likewise, constraining more attributes does not consistently lower performance (C1 \texttt{4prop}: 0.8 vs.\ \texttt{2prop}: 0.4 for \gpt{}).
The open models succeed on only a few templates, with performance never above 0.4.

%% file: figures/qualitative-example/qualitative_examples.tex
\begin{figure*}[t]
\centering
\footnotesize
\setlength{\tabcolsep}{2pt}
\setlength{\fboxsep}{0pt}
\setlength{\fboxrule}{0.4pt}
% one image: fixed height, top-aligned, thin grey frame
\newcommand{\qimg}[1]{\raisebox{-\height}{\fcolorbox{black!30}{white}{\includegraphics[width=0.97\linewidth,height=2.3cm,keepaspectratio]{qualitative-example/#1}}}}
\newcommand{\qok}[1][]{\textcolor{green!50!black}{\ding{51}\ #1}}
\newcommand{\qbad}[1]{\textcolor{red!70!black}{\ding{55}\ #1}}
\newcommand{\qhead}[1]{\multicolumn{4}{@{}l}{\textbf{#1}}\\[1pt]}
\newcommand{\qprompt}[1]{\multicolumn{4}{@{}p{\textwidth}}{\raggedright\textcolor{black!80}{Prompt: #1}}\\[-7pt]}
% gap between examples: grey dashed line with some space above and below
\newcommand{\qgap}{\\[5pt]\arrayrulecolor{black!35}\cdashline{1-4}[2pt/2pt]\arrayrulecolor{black}\noalign{\vskip 4pt}}
\begin{tabular}{@{}*{4}{>{\centering\arraybackslash}p{0.245\textwidth}}@{}}
\textbf{GPT-Image-2} & \textbf{Gemini-3-Pro-Image} & \textbf{FLUX.1-dev} & \textbf{Qwen-Image} \\[4pt]
\hline\\
%
% ---- row 1 -------------------------------------------------------------------------
\qhead{1SAT: C1\_1prop $\cdot$ 5 objects}
\qprompt{A gray object is in the scene.}
\qimg{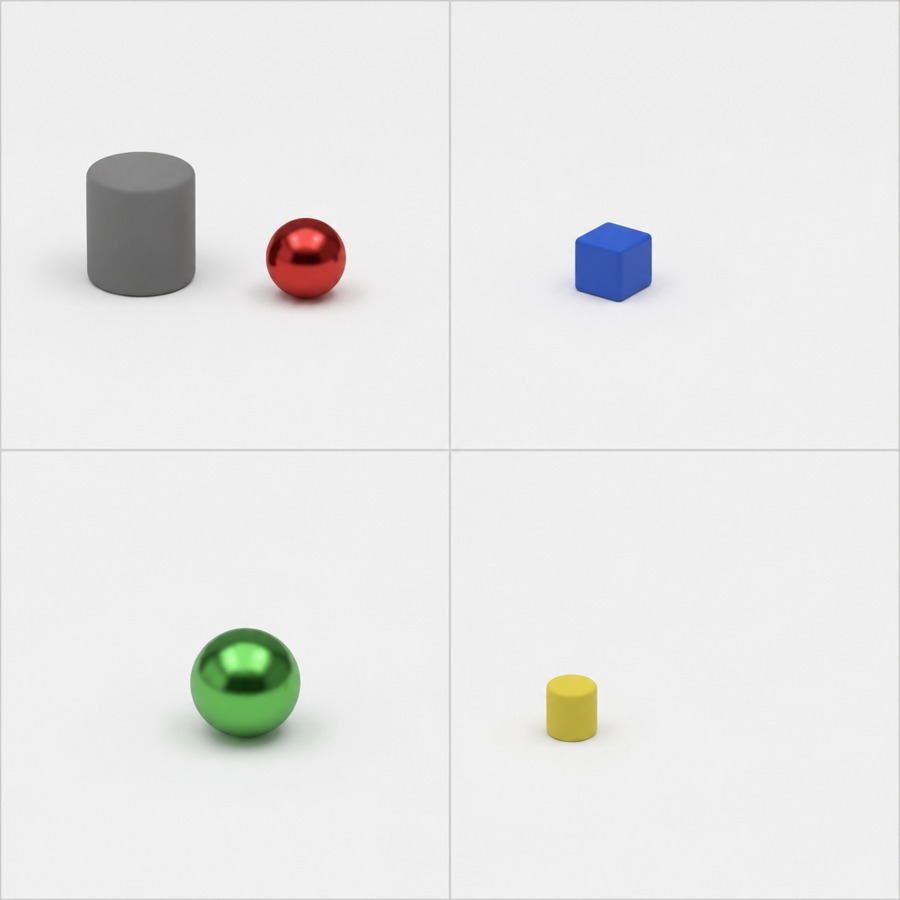} & \qimg{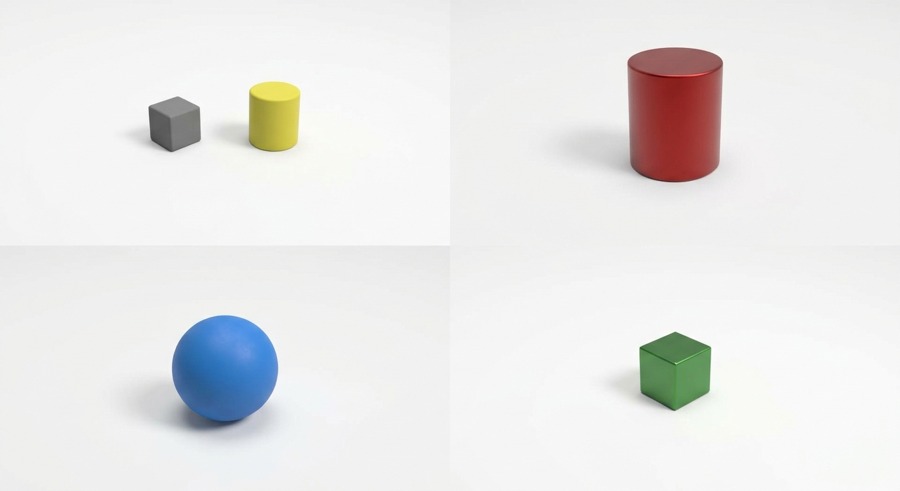} & \qimg{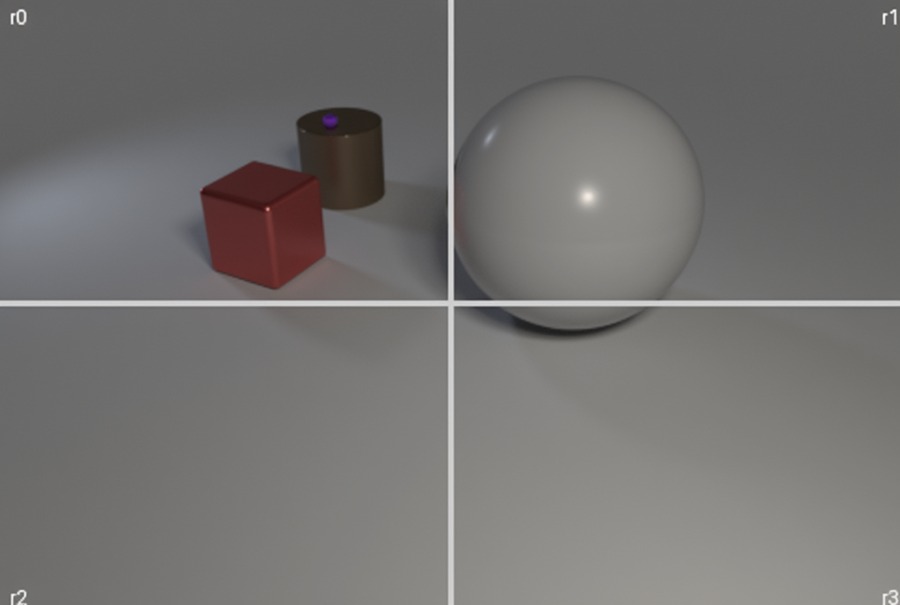} & \qimg{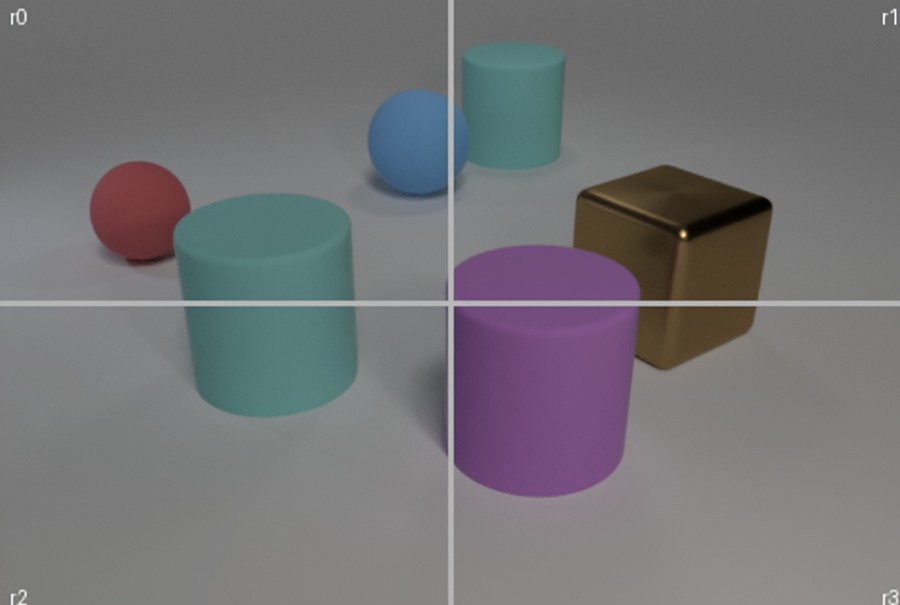} \\[2pt]
\qok & \qok & \qbad{3 objects (5 requested)} & \qbad{6 objects (5 requested)} \qgap
\\
%
% ---- row 2 -------------------------------------------------------------------------
\qhead{1SAT: C5\_pair\_propA\_relC $\cdot$ 6 objects}
\qprompt{If there is a metal object and a cylinder object, then the former must be in front of the latter. There must be at least one such pair.}
\qimg{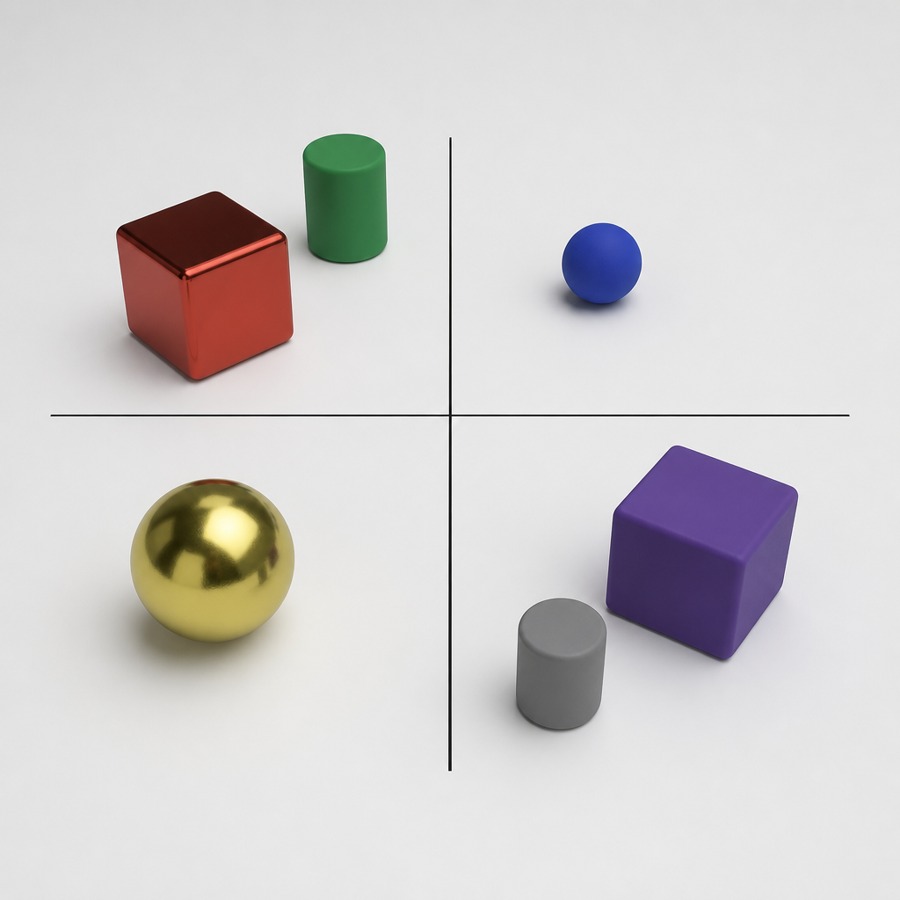} & \qimg{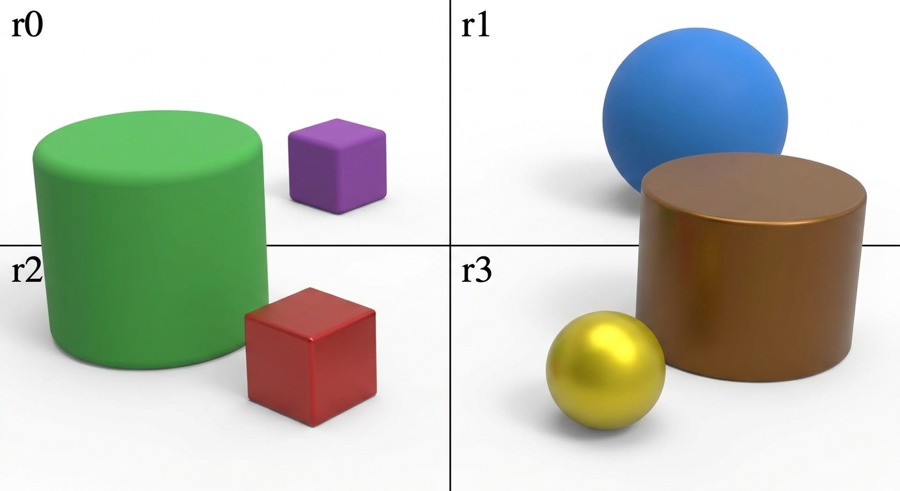} & \qimg{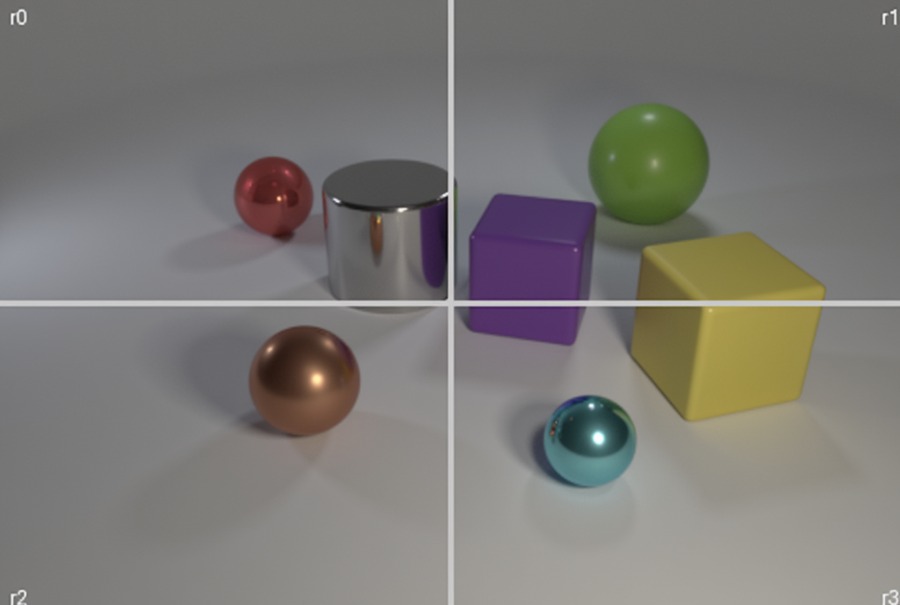} & \qimg{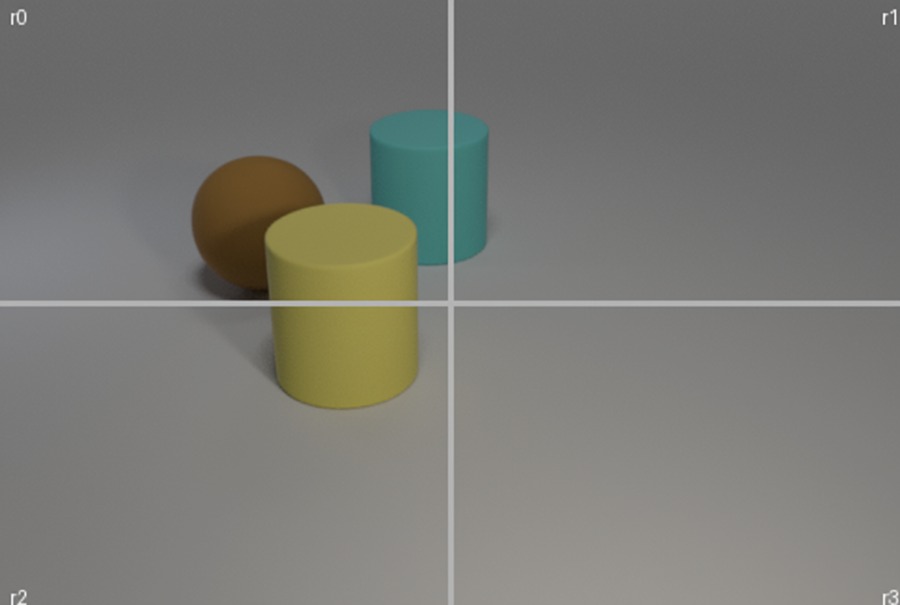} \\[2pt]
\qbad{constraint violated} & \qok & \qbad{7 objects (6 requested)} & \qbad{3 objects (6 requested)} \qgap
\\
%
% ---- row 3 -------------------------------------------------------------------------
\qhead{1SAT: C8\_1prop\_atmost $\cdot$ 5 objects}
\qprompt{There is at most one cylinder object.}
\qimg{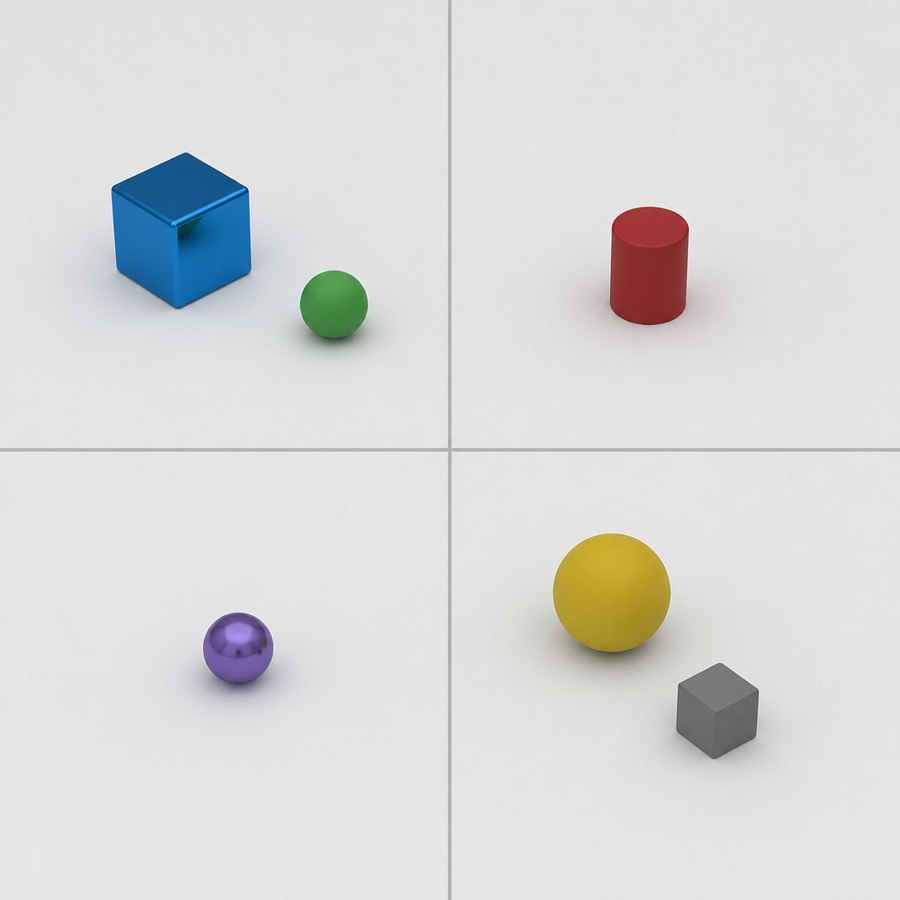} & \qimg{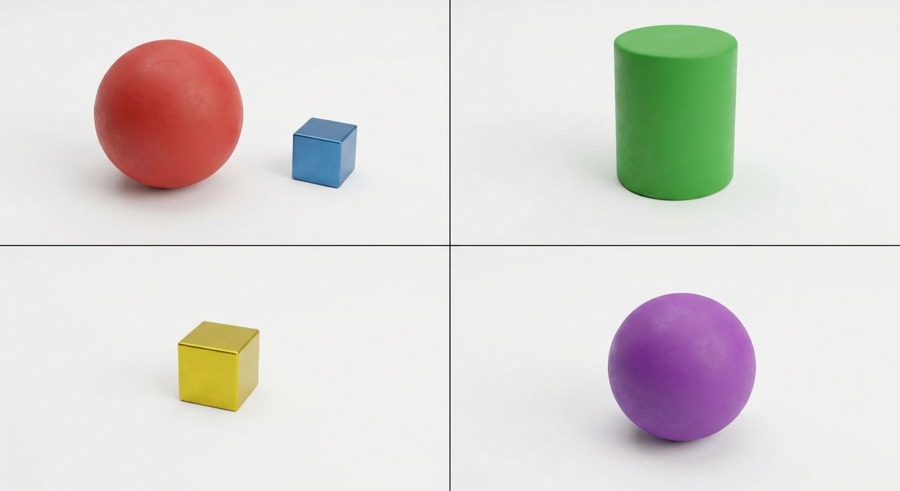} & \qimg{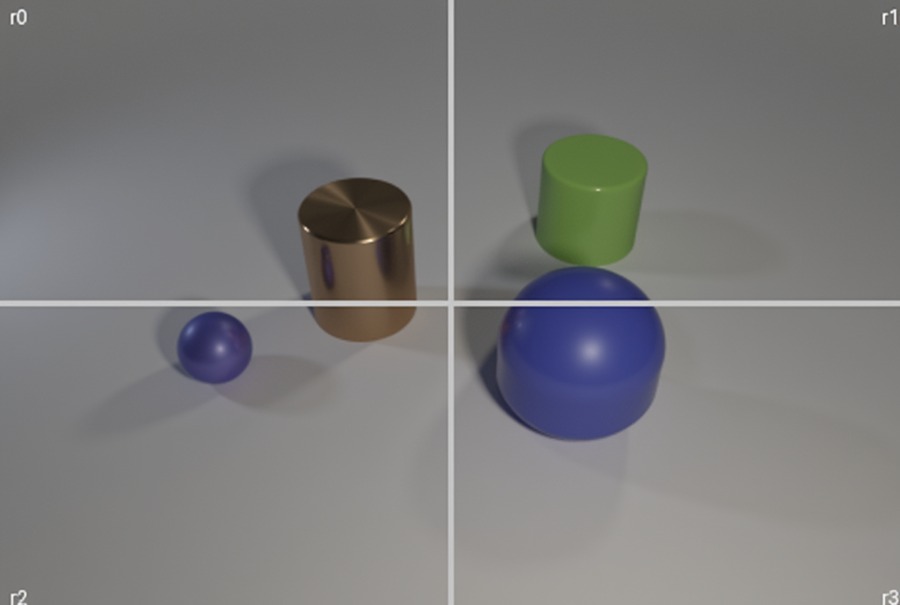} & \qimg{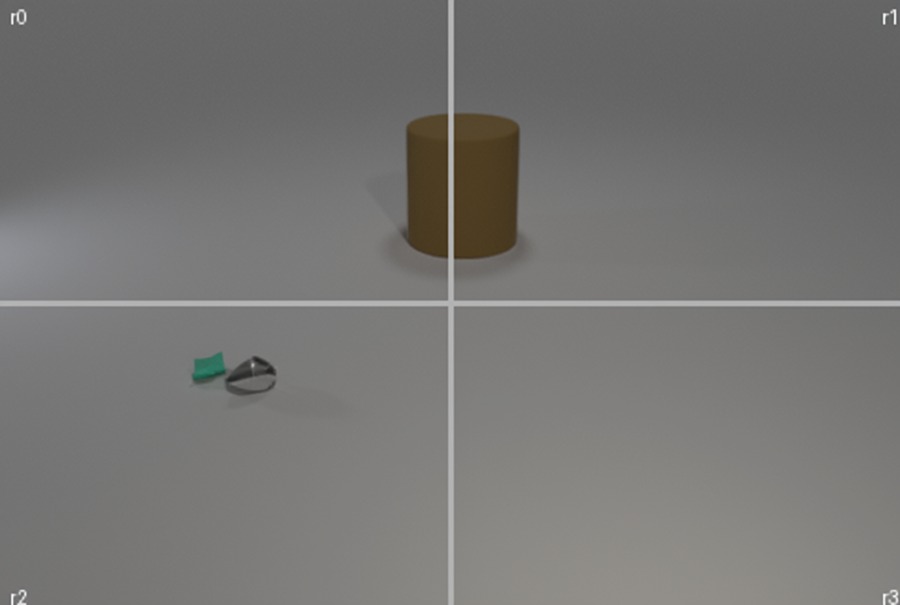} \\[2pt]
\qbad{6 objects (5 requested)} & \qok & \qbad{4 objects (5 requested)} & \qbad{3 objects, 2 deformed\newline(5 requested)} \qgap
\\
%
% ---- row 4 -------------------------------------------------------------------------
\qhead{3SAT: C8\_2prop\_atmost, C1\_1prop\_2val\_neg, C3\_3propA\_1propC $\cdot$ 3 objects}
\qprompt{\newline
(1) There is at most 1 object that is both in r1 and yellow.\newline
(2) Some object is neither cube nor sphere.\newline
(3) At least one object is sphere, metal, and in r3, and every such object must be blue.}
\qimg{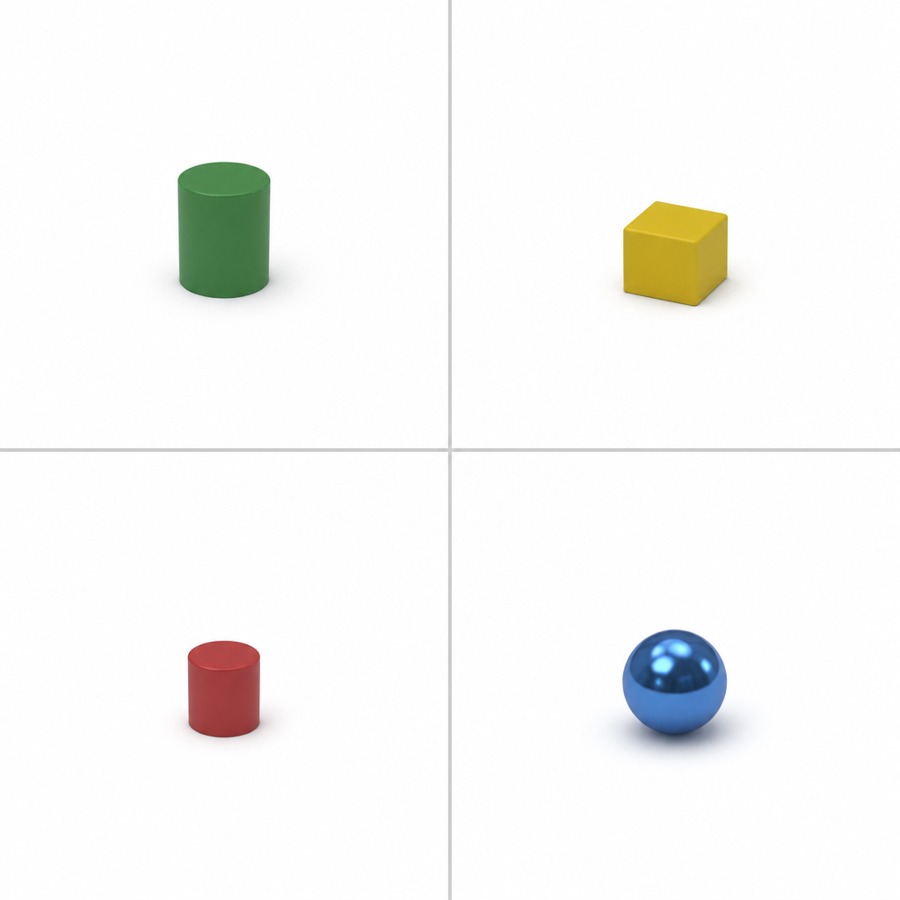} & \qimg{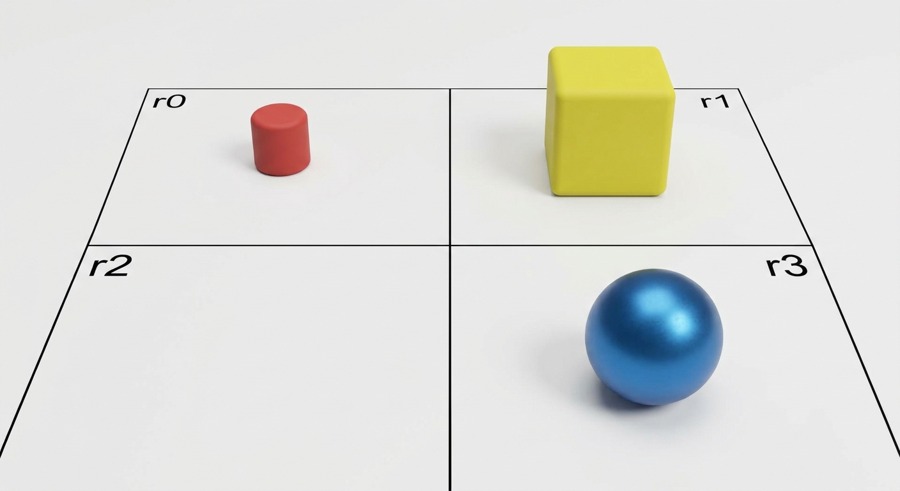} & \qimg{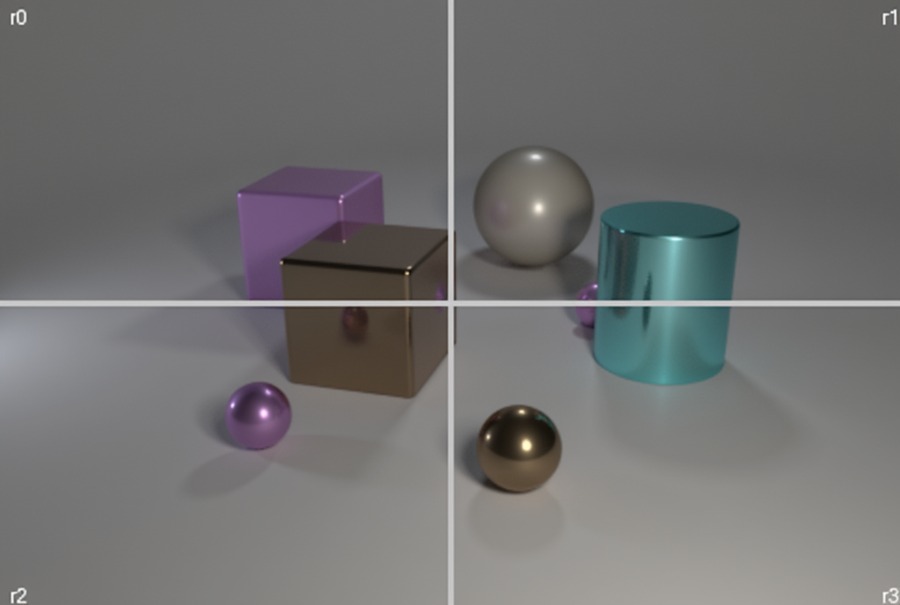} & \qimg{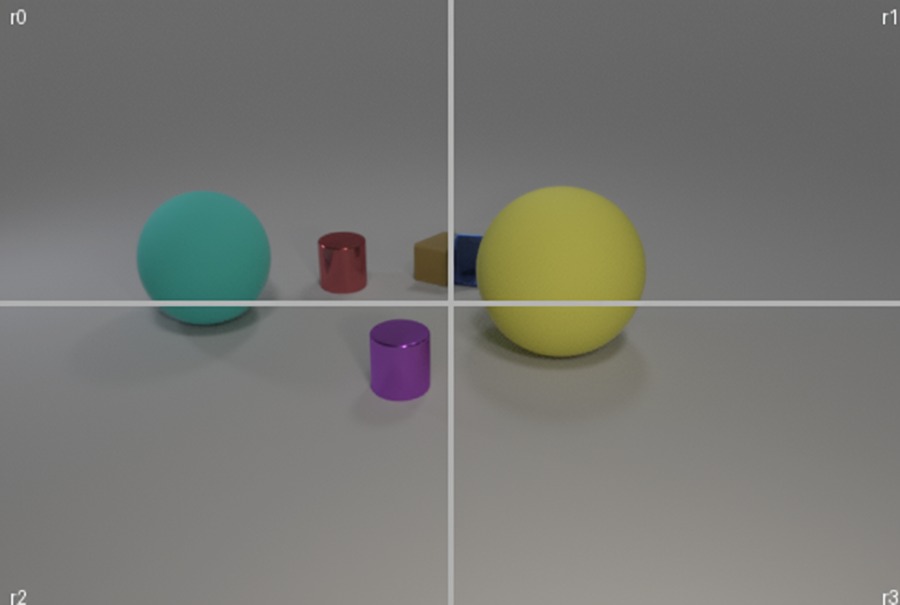} \\[2pt]
\qbad{4 objects (3 requested)} & \qok & \qbad{7 objects (3 requested)} & \qbad{6 objects (3 requested)} \qgap
\\
%
% ---- row 5 -------------------------------------------------------------------------
\qhead{1UNSAT: C2\_1prop\_2val\_neg $\cdot$ 5 objects}
\qprompt{No object is either rubber or metal.}
\qimg{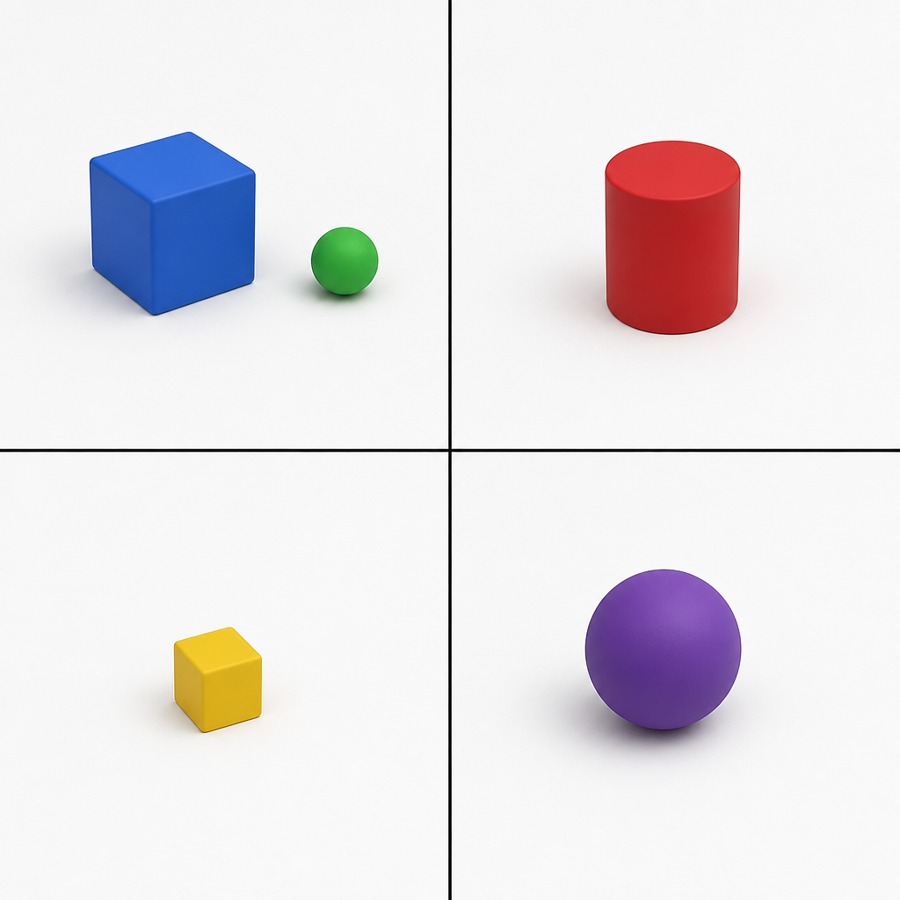} & \qimg{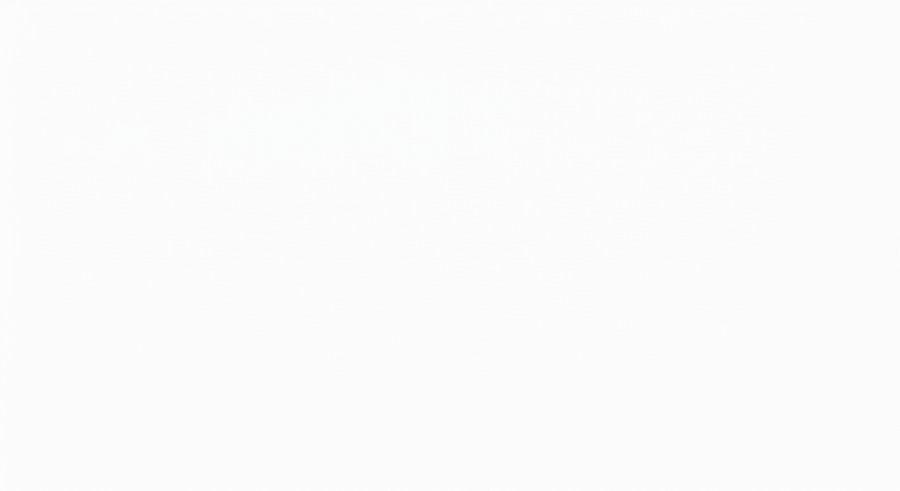} & \qimg{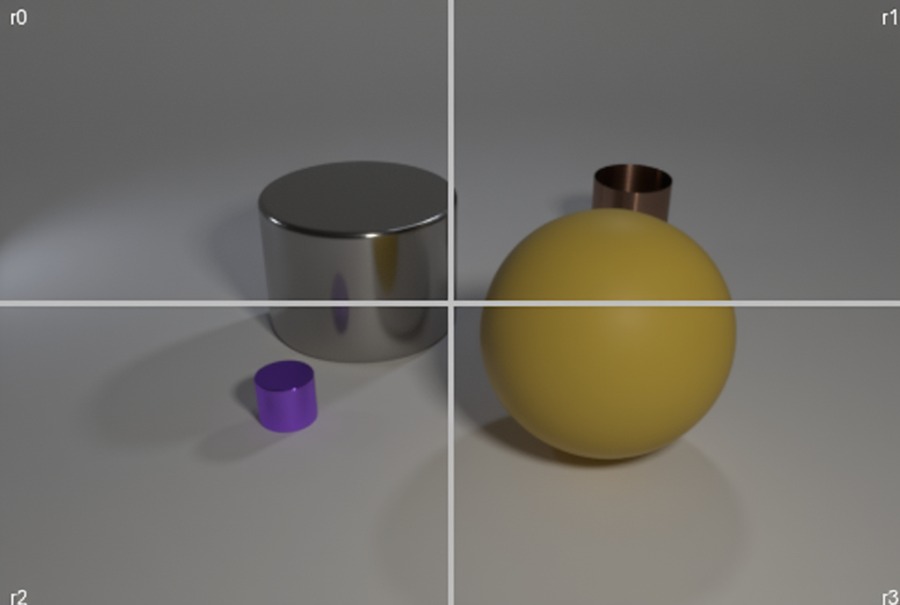} & \qimg{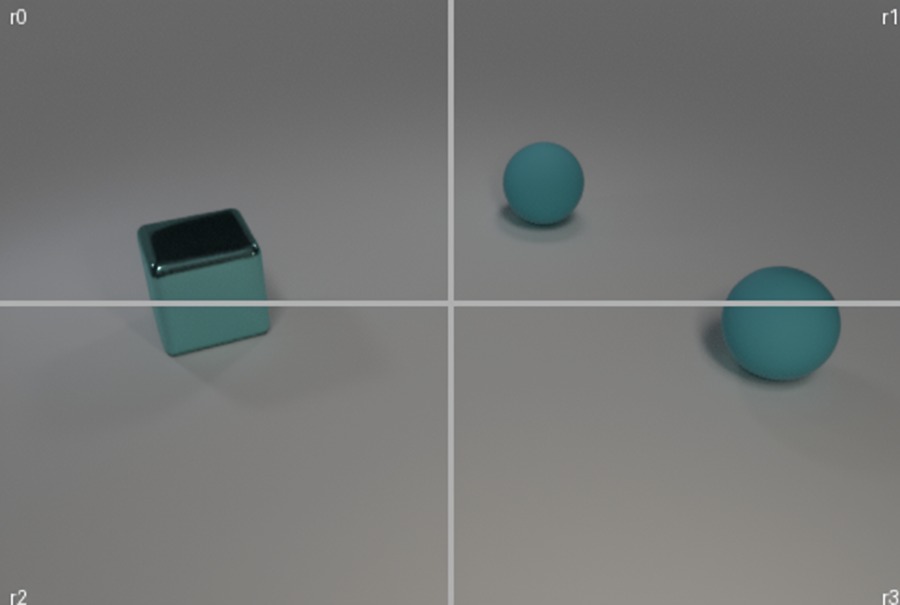} \\[2pt]
\qbad{objects rendered} & \qok[empty scene] & \qbad{objects rendered} & \qbad{objects rendered} \\
\end{tabular}
\caption{Images generated by each model for the same prompt (short version), with the human verdict (\ding{51}/\ding{55}) and, for failures, the reason.}
\label{fig:qualitative}
\end{figure*}

%% file: main-preprint.bbl
\begin{thebibliography}{37}
\providecommand{\natexlab}[1]{#1}

\bibitem[{Abraham et~al.(2024)Abraham, Alirezaie, and De~Raedt}]{abraham2024clevrpoc}
Savitha~Sam Abraham, Marjan Alirezaie, and Luc De~Raedt. 2024.
\newblock {CLEVR-POC}: Reasoning-intensive visual question answering in partially observable environments.
\newblock In \emph{Proceedings of the 2024 Joint International Conference on Computational Linguistics, Language Resources and Evaluation (LREC-COLING 2024)}.

\bibitem[{{Black Forest Labs}(2024)}]{bfl2024flux}
{Black Forest Labs}. 2024.
\newblock {FLUX.1}.
\newblock \url{https://github.com/black-forest-labs/flux}.

\bibitem[{Chen et~al.(2025)Chen, Lin, Xu, Shen, Yao, Rimchala, Zhang, and Huang}]{chen2025r2ibench}
Kaijie Chen, Zihao Lin, Zhiyang Xu, Ying Shen, Yuguang Yao, Joy Rimchala, Jiaxin Zhang, and Lifu Huang. 2025.
\newblock \href {https://doi.org/10.18653/v1/2025.emnlp-main.636} {{R}2{I}-{B}ench: Benchmarking reasoning-driven text-to-image generation}.
\newblock In \emph{Proceedings of the 2025 Conference on Empirical Methods in Natural Language Processing}, pages 12595--12630, Suzhou, China. Association for Computational Linguistics.

\bibitem[{Chen et~al.(2018)Chen, Rubanova, Bettencourt, and Duvenaud}]{chen2018neural}
Ricky T.~Q. Chen, Yulia Rubanova, Jesse Bettencourt, and David~K Duvenaud. 2018.
\newblock \href {https://nips.cc} {Neural ordinary differential equations}.
\newblock In \emph{Advances in Neural Information Processing Systems (NeurIPS)}, volume~31, pages 6571--6583.

\bibitem[{Eiter et~al.(2009)Eiter, Ianni, and Krennwallner}]{eiter2009answer}
Thomas Eiter, Giovambattista Ianni, and Thomas Krennwallner. 2009.
\newblock Answer set programming: A primer.
\newblock In \emph{Reasoning Web International Summer School}, pages 40--110. Springer.

\bibitem[{Elfatairy et~al.(2026)Elfatairy, Bravo, Bader, and Akata}]{elfatairy2026negt2ibench}
Omar Elfatairy, Maria~A. Bravo, Jessica Bader, and Zeynep Akata. 2026.
\newblock {NegT2IBench}: When negation changes the picture. a polarity benchmark for text-to-image models.
\newblock \emph{arXiv preprint arXiv:2610.03084}.

\bibitem[{Gebser et~al.(2019)Gebser, Kaminski, Kaufmann, and Schaub}]{gebser2019clingo}
Martin Gebser, Roland Kaminski, Benjamin Kaufmann, and Torsten Schaub. 2019.
\newblock Multi-shot {ASP} solving with clingo.
\newblock \emph{Theory and Practice of Logic Programming}, 19(1):27--82.

\bibitem[{Ghosh et~al.(2023)Ghosh, Hajishirzi, and Schmidt}]{ghosh2023geneval}
Dhruba Ghosh, Hannaneh Hajishirzi, and Ludwig Schmidt. 2023.
\newblock {GenEval}: An object-focused framework for evaluating text-to-image alignment.
\newblock In \emph{Advances in Neural Information Processing Systems (NeurIPS), Datasets and Benchmarks Track}.

\bibitem[{Goodfellow et~al.(2014)Goodfellow, Pouget-Abadie, Mirza, Xu, Warde-Farley, Ozair, Courville, and Bengio}]{goodfellow2014generative}
Ian~J. Goodfellow, Jean Pouget-Abadie, Mehdi Mirza, Bing Xu, David Warde-Farley, Sherjil Ozair, Aaron~C. Courville, and Yoshua Bengio. 2014.
\newblock \href {https://proceedings.neurips.cc/paper_files/paper/2014/file/f033ed80deb0234979a61f95710dbe25-Paper.pdf} {Generative adversarial nets}.
\newblock In \emph{Advances in Neural Information Processing Systems}, volume~27, pages 2672--2680. Curran Associates, Inc.

\bibitem[{{Google DeepMind}(2025)}]{google2025geminiimage}
{Google DeepMind}. 2025.
\newblock {Gemini 3 Pro Image}.
\newblock \url{https://ai.google.dev/gemini-api/docs/models}.

\bibitem[{Hessel et~al.(2021)Hessel, Holtzman, Forbes, Le~Bras, and Choi}]{hessel2021clipscore}
Jack Hessel, Ari Holtzman, Maxwell Forbes, Ronan Le~Bras, and Yejin Choi. 2021.
\newblock {CLIPScore}: A reference-free evaluation metric for image captioning.
\newblock In \emph{Proceedings of the 2021 Conference on Empirical Methods in Natural Language Processing (EMNLP)}, pages 7514--7528.

\bibitem[{Ho et~al.(2020)Ho, Jain, and Abbeel}]{ho2020denoising}
Jonathan Ho, Ajay Jain, and Pieter Abbeel. 2020.
\newblock Denoising diffusion probabilistic models.
\newblock In \emph{Advances in Neural Information Processing Systems}, volume~33, pages 6840--6851. Curran Associates, Inc.

\bibitem[{Hu et~al.(2022)Hu, Shen, Wallis, Allen-Zhu, Li, Wang, Wang, and Chen}]{hu2022lora}
Edward~J. Hu, Yelong Shen, Phillip Wallis, Zeyuan Allen-Zhu, Yuanzhi Li, Shean Wang, Lu~Wang, and Weizhu Chen. 2022.
\newblock {LoRA}: Low-rank adaptation of large language models.
\newblock In \emph{International Conference on Learning Representations (ICLR)}.

\bibitem[{Huang et~al.(2025)Huang, Duan, Sun, Xie, Li, and Liu}]{huang2025t2i}
Kaiyi Huang, Chengqi Duan, Kaiyue Sun, Enze Xie, Zhenguo Li, and Xihui Liu. 2025.
\newblock T2i-compbench++: An enhanced and comprehensive benchmark for compositional text-to-image generation.
\newblock \emph{IEEE Transactions on Pattern Analysis and Machine Intelligence}, 47(5):3563--3579.

\bibitem[{Huang et~al.(2023)Huang, Sun, Xie, Li, and Liu}]{huang2023t2icompbench}
Kaiyi Huang, Kaiyue Sun, Enze Xie, Zhenguo Li, and Xihui Liu. 2023.
\newblock {T2I-CompBench}: A comprehensive benchmark for open-world compositional text-to-image generation.
\newblock In \emph{Advances in Neural Information Processing Systems (NeurIPS), Datasets and Benchmarks Track}.

\bibitem[{Hurst et~al.(2024)Hurst, Lerer, Goucher, Perelman, Ramesh, Clark, Ostrow, Welihinda, Hayes, Radford et~al.}]{hurst2024gpt}
Aaron Hurst, Adam Lerer, Adam~P Goucher, Adam Perelman, Aditya Ramesh, Aidan Clark, AJ~Ostrow, Akila Welihinda, Alan Hayes, Alec Radford, et~al. 2024.
\newblock Gpt-4o system card.
\newblock \emph{arXiv preprint arXiv:2410.21276}.

\bibitem[{Johnson et~al.(2017)Johnson, Hariharan, van~der Maaten, Fei-Fei, Zitnick, and Girshick}]{johnson2017clevr}
Justin Johnson, Bharath Hariharan, Laurens van~der Maaten, Li~Fei-Fei, C.~Lawrence Zitnick, and Ross Girshick. 2017.
\newblock {CLEVR}: A diagnostic dataset for compositional language and elementary visual reasoning.
\newblock In \emph{Proceedings of the IEEE Conference on Computer Vision and Pattern Recognition (CVPR)}, pages 2901--2910.

\bibitem[{Kaji{\'c} et~al.(2024)Kaji{\'c}, Wiles, Albuquerque, Bauer, Wang, Pont-Tuset, and Nematzadeh}]{kajic2024geckonum}
Ivana Kaji{\'c}, Olivia Wiles, Isabela Albuquerque, Matthias Bauer, Su~Wang, Jordi Pont-Tuset, and Aida Nematzadeh. 2024.
\newblock Evaluating numerical reasoning in text-to-image models.
\newblock \emph{arXiv preprint arXiv:2406.14774}.

\bibitem[{Kamath et~al.(2025)Kamath, Chang, Krishna, Zettlemoyer, Hu, and Ghazvininejad}]{kamath2025geneval}
Amita Kamath, Kai-Wei Chang, Ranjay Krishna, Luke Zettlemoyer, Yushi Hu, and Marjan Ghazvininejad. 2025.
\newblock Geneval 2: Addressing benchmark drift in text-to-image evaluation.
\newblock \emph{arXiv preprint arXiv:2512.16853}.

\bibitem[{Lee et~al.(2024)Lee, Kim, Park, Kim, and Seo}]{lee2024prometheusvision}
Seongyun Lee, Seungone Kim, Sue~Hyun Park, Geewook Kim, and Minjoon Seo. 2024.
\newblock \href {https://aclanthology.org/2024.findings-acl.672/} {Prometheus-vision: Vision-language model as a judge for fine-grained evaluation}.
\newblock In \emph{Findings of the Association for Computational Linguistics: ACL 2024}.

\bibitem[{Li et~al.(2024)Li, Lin, Pathak, Li, Fei, Wu, Ling, Xia, Zhang, Neubig, and Ramanan}]{li2024genaibench}
Baiqi Li, Zhiqiu Lin, Deepak Pathak, Jiayao Li, Yixin Fei, Kewen Wu, Tiffany Ling, Xide Xia, Pengchuan Zhang, Graham Neubig, and Deva Ramanan. 2024.
\newblock {GenAI-Bench}: Evaluating and improving compositional text-to-visual generation.
\newblock \emph{arXiv preprint arXiv:2406.13743}.

\bibitem[{Li et~al.(2026)Li, Qiu, Koo, Han, and Bouyarmane}]{li2026dit}
Qi~Li, Shuwen Qiu, Kee~Kiat Koo, Julien Han, and Karim Bouyarmane. 2026.
\newblock Dit-vton: Diffusion transformer framework for unified multi-category virtual try-on and virtual try-all with integrated image editing.
\newblock In \emph{2026 IEEE/CVF Winter Conference on Applications of Computer Vision (WACV)}, pages 202--211. IEEE.

\bibitem[{Lin et~al.(2024)Lin, Pathak, Li, Li, Xia, Neubig, Zhang, and Ramanan}]{lin2024vqascore}
Zhiqiu Lin, Deepak Pathak, Baiqi Li, Jiayao Li, Xide Xia, Graham Neubig, Pengchuan Zhang, and Deva Ramanan. 2024.
\newblock Evaluating text-to-visual generation with image-to-text generation.
\newblock In \emph{European Conference on Computer Vision}, pages 366--384. Springer.

\bibitem[{Lipman et~al.(2023)Lipman, Chen, Ben-Hamu, Nickel, and Le}]{lipman2023flow}
Yaron Lipman, Ricky T.~Q. Chen, Heli Ben-Hamu, Maximilian Nickel, and Matt Le. 2023.
\newblock \href {https://openreview.net/forum?id=PqvMRDCJT9t} {Flow matching for generative modeling}.
\newblock In \emph{International Conference on Learning Representations (ICLR)}.

\bibitem[{Ni(2025)}]{ni2025ai}
Jun Ni. 2025.
\newblock An ai toolchain framework for optimizing the architectural design process.
\newblock \emph{Journal of Frontiers in Science and Engineering}, 1(9):15--20.

\bibitem[{{OpenAI}(2026)}]{openai2026gptimage2}
{OpenAI}. 2026.
\newblock {GPT-Image-2}.
\newblock \url{https://platform.openai.com/docs/models}.

\bibitem[{Radford et~al.(2021)Radford, Kim, Hallacy, Ramesh, Goh, Agarwal, Sastry, Askell, Mishkin, Clark, Krueger, and Sutskever}]{radford2021clip}
Alec Radford, Jong~Wook Kim, Chris Hallacy, Aditya Ramesh, Gabriel Goh, Sandhini Agarwal, Girish Sastry, Amanda Askell, Pamela Mishkin, Jack Clark, Gretchen Krueger, and Ilya Sutskever. 2021.
\newblock Learning transferable visual models from natural language supervision.
\newblock In \emph{Proceedings of the 38th International Conference on Machine Learning (ICML)}.

\bibitem[{Ren et~al.(2024)Ren, Liu, Zeng, Lin, Li, Cao, Chen, Huang, Chen, Yan et~al.}]{ren2024grounded}
Tianhe Ren, Shilong Liu, Ailing Zeng, Jing Lin, Kunchang Li, He~Cao, Jiayu Chen, Xinyu Huang, Yukang Chen, Feng Yan, et~al. 2024.
\newblock Grounded sam: Assembling open-world models for diverse visual tasks.
\newblock \emph{arXiv preprint arXiv:2401.14159}.

\bibitem[{Rezende and Mohamed(2015)}]{rezende2015variational}
Danilo~Jimenez Rezende and Shakir Mohamed. 2015.
\newblock \href {https://proceedings.mlr.press/v37/rezende15.html} {Variational inference with normalizing flows}.
\newblock In \emph{Proceedings of the 32nd International Conference on Machine Learning (ICML)}, volume~37 of \emph{JMLR Workshop and Conference Proceedings}, pages 1530--1538, Lille, France. JMLR.org.

\bibitem[{Soldani(2026)}]{soldani2026claude}
David Soldani. 2026.
\newblock Claude code complete user handbook.
\newblock \emph{arXiv preprint arXiv:2608.26742}.

\bibitem[{Song et~al.(2020)Song, Meng, and Ermon}]{song2020denoising}
Jiaming Song, Chenlin Meng, and Stefano Ermon. 2020.
\newblock Denoising diffusion implicit models.
\newblock \emph{arXiv preprint arXiv:2010.02502}.

\bibitem[{Sun et~al.(2025)Sun, Fang, Duan, Liu, and Liu}]{sun2025t2ireasonbench}
Kaiyue Sun, Rongyao Fang, Chengqi Duan, Xian Liu, and Xihui Liu. 2025.
\newblock {T2I-ReasonBench}: Benchmarking reasoning-informed text-to-image generation.
\newblock \emph{arXiv preprint arXiv:2508.17472}.

\bibitem[{Temsah et~al.(2024)Temsah, Alhuzaimi, Almansour, Aljamaan, Alhasan, Batarfi, Altamimi, Alharbi, Alsuhaibani, Alwakeel et~al.}]{temsah2024art}
Mohamad-Hani Temsah, Abdullah~N Alhuzaimi, Mohammed Almansour, Fadi Aljamaan, Khalid Alhasan, Munirah~A Batarfi, Ibraheem Altamimi, Amani Alharbi, Adel~Abdulaziz Alsuhaibani, Leena Alwakeel, et~al. 2024.
\newblock Art or artifact: evaluating the accuracy, appeal, and educational value of ai-generated imagery in dall{\textperiodcentered} e 3 for illustrating congenital heart diseases.
\newblock \emph{Journal of Medical Systems}, 48(1):54.

\bibitem[{van~den Oord et~al.(2017)van~den Oord, Vinyals, and Kavukcuoglu}]{van2017neural}
Aaron van~den Oord, Oriol Vinyals, and Koray Kavukcuoglu. 2017.
\newblock \href {https://papers.nips.cc/paper/7210-neural-discrete-representation-learning} {Neural discrete representation learning}.
\newblock In \emph{Advances in Neural Information Processing Systems (NIPS)}, volume~30, pages 6306--6315. Curran Associates, Inc.

\bibitem[{Wu et~al.(2025)Wu, Li, Zhou, Lin, Gao, Yan, Yin, Bai, Xu, Chen et~al.}]{wu2025qwen}
Chenfei Wu, Jiahao Li, Jingren Zhou, Junyang Lin, Kaiyuan Gao, Kun Yan, Sheng-ming Yin, Shuai Bai, Xiao Xu, Yilei Chen, et~al. 2025.
\newblock Qwen-image technical report.
\newblock \emph{arXiv preprint arXiv:2508.02324}.

\bibitem[{Zhang et~al.(2024)Zhang, Zhang, Dong, Zang, and Wang}]{zhang2024longclip}
Beichen Zhang, Pan Zhang, Xiaoyi Dong, Yuhang Zang, and Jiaqi Wang. 2024.
\newblock Long-{CLIP}: Unlocking the long-text capability of {CLIP}.
\newblock In \emph{European Conference on Computer Vision (ECCV)}.

\bibitem[{Zhang et~al.(2025)Zhang, Jiang, Xu, Chen, Jin, Lu, Zhang, Yong, Luo, and Luo}]{zhang2025worldgenbench}
Daoan Zhang, Che Jiang, Ruoshi Xu, Biaoxiang Chen, Zijian Jin, Yutian Lu, Jianguo Zhang, Liang Yong, Jiebo Luo, and Shengda Luo. 2025.
\newblock Worldgenbench: A world-knowledge-integrated benchmark for reasoning-driven text-to-image generation.
\newblock \emph{arXiv preprint arXiv:2505.01490}.

\end{thebibliography}
